\documentclass[sigconf]{acmart}

\copyrightyear{2026}
\acmYear{2026}
\setcopyright{cc}
\setcctype{by-nc-nd}
\acmConference[MM '26]{Proceedings of the 34th ACM International Conference on Multimedia}{November 10--14, 2026}{Rio de Janeiro, Brazil}
\acmBooktitle{Proceedings of the 34th ACM International Conference on Multimedia (MM '26), November 10--14, 2026, Rio de Janeiro, Brazil}
\acmDOI{10.1145/3767308.3835002}
\acmISBN{979-8-4007-2213-4/2026/11} 

\usepackage{graphicx}  
\usepackage{multibib}

\usepackage{caption}
\usepackage{subcaption}
\usepackage{multirow}  
\usepackage{amsthm}

\usepackage{mathalpha}
\usepackage{bbding}
\usepackage{url}            %
\usepackage[table]{xcolor}

\newlength\savewidth\newcommand\shline{\noalign{\global\savewidth\arrayrulewidth \global\arrayrulewidth 1pt}\hline\noalign{\global\arrayrulewidth\savewidth}}
\newcommand{\tablestyle}[2]{\setlength{\tabcolsep}{#1}\renewcommand{\arraystretch}{#2}\centering\footnotesize}

\newcolumntype{x}[1]{>{\centering\arraybackslash}p{#1pt}}
\newcolumntype{y}[1]{>{\raggedright\arraybackslash}p{#1pt}}
\newcolumntype{z}[1]{>{\raggedleft\arraybackslash}p{#1pt}}

\newcommand{\app}{\raise.17ex\hbox{$\scriptstyle\sim$}}

\definecolor{baselinecolor}{gray}{.9}
\newcommand{\baseline}[1]{\cellcolor{baselinecolor}{#1}}
 
\newcolumntype{^}{>{\currentrowstyle}}

\definecolor{dt}{gray}{0.7}  %

\usepackage[capitalize]{cleveref}
\crefname{section}{Sec.}{Secs.}
\Crefname{section}{Section}{Sections}
\Crefname{table}{Table}{Tables}
\crefname{table}{Tab.}{Tabs.}

\newcolumntype{S}{@{}>{\lrbox0}l<{\endlrbox}}  %
\definecolor{lightgreen}{HTML}{D8ECD1}
\newcommand{\better}[1]{\colorbox{lightgreen}{#1}}

\newcommand{\ie}{{\emph{i.e.}}, }

\newcommand{\eg}{{\emph{e.g.}}, }

\begin{document}
 
\title{SignLlama: Enhancing Gloss-free Sign Language Translation by Prioritizing Visual Features for LLMs}

\author{Shiwei Gan}
\authornote{Both authors contributed equally to this research.} 
\orcid{0000-0003-3360-4321} 
\affiliation{%
  \institution{State Key Laboratory of Novel Software Technology, Nanjing University}
  \city{Nanjing}
  \state{Jiangsu}
  \country{China}}
\email{sw@nju.edu.cn} 
 
\author{Xiao Liu}
\orcid{0000-0001-6943-9861} 
\authornotemark[1]
\affiliation{%
  \institution{State Key Laboratory of Novel Software Technology, Nanjing University}
  \city{Suzhou}
  \state{Jiangsu}
    \country{China}
}
\email{liuxiaox@smail.nju.edu.cn}

\author{Yafeng Yin}
\orcid{0000-0002-9497-6244} 
\authornote{Yafeng Yin is the corresponding author.} 
\affiliation{
  \institution{State Key Laboratory of Novel Software Technology, Nanjing University}
  \city{Suzhou}
  \state{Jiangsu}
    \country{China}
}
\email{ yafeng@nju.edu.cn}

\author{Zhiwei Jiang}
\orcid{0000-0001-5243-4992}  
\affiliation{
  \institution{State Key Laboratory of Novel Software Technology, Nanjing University}
  \city{Suzhou}
  \state{Jiangsu}
    \country{China}
}
\email{jzw@nju.edu}
\author{Bowen Guo}
\orcid{0009-0000-6390-4398} 
\affiliation{
  \institution{State Key Laboratory of Novel Software Technology, Nanjing University}
  \city{Suzhou}
  \state{Jiangsu}
    \country{China}
}  
\email{602024720001@smail.nju.edu.cn}

\author{Lei Xie}
\orcid{0000-0002-2994-6743} 
\affiliation{%
  \institution{State Key Laboratory of Novel Software Technology, Nanjing University}
  \city{Nanjing}
  \state{Jiangsu}
  \country{China}
} 
\email{lxie@nju.edu.cn}

\author{Sanglu Lu} 
\orcid{0000-0003-1467-4519}
\affiliation{%
  \institution{State Key Laboratory of Novel Software Technology, Nanjing University}
  \city{Nanjing}
  \state{Jiangsu}
  \country{China} } 
\email{sanglu@nju.edu.cn}

\author{Hongkai Wen}
\orcid{0000-0003-1159-090X}
\affiliation{%
  \institution{University of Warwick}
  \city{Coventry}
  \country{United Kingdom}}
\email{hongkai.wen@warwick.ac.uk}

\renewcommand{\shortauthors}{Shiwei Gan et al.}
 
\begin{abstract}
Large Language Models (LLMs) have achieved remarkable success across a wide range of tasks. However, fine-tuning LLMs for Gloss-Free Sign Language Translation (GFSLT) remains a  challenge. In this paper, we investigate how to effectively adapt LLMs to the GFSLT task. We show that there are two key issues that need to be solved: (1) the inherent distributional gap between visual feature inputs and text feature inputs makes it difficult for  LLMs to interpret visual inputs; and (2) existing approaches typically concatenate visual and textual features in an autoregressive framework, which leads to the model overemphasizing textual inputs and deprioritizing visual cues, as LLMs are pretrained predominantly on text-centric data.   To address the first challenge, we propose a simple yet effective method named Filtered  Pseudo-Gloss CTC Pretraining, which leverages filtered pseudo-gloss sequences generated from text sequences to supervise the training of the visual backbone. To tackle the second issue, we introduce a Visual-Prioritized Distillation training strategy. Specifically, we define a visual-only prediction path in which text inputs are masked, and the model is required to generate the target sequence relying solely on visual inputs.  To guide this path, the outputs from the standard visual-textual prediction are then distilled into the visual-only prediction path, encouraging the model to prioritize visual features. Comprehensive experiments and qualitative analyses demonstrate the effectiveness of the proposed model. The proposed SignLlama achieves very competitive performance on multiple datasets for GFSLT tasks, without using any extra modalities or external sign language datasets for pretraining.  

\end{abstract}

\begin{CCSXML}
<ccs2012>
   <concept>
       <concept_id>10010147.10010178.10010224.10010225.10010228</concept_id>
       <concept_desc>Computing methodologies~Activity recognition and understanding</concept_desc>
       <concept_significance>500</concept_significance>
       </concept>
   <concept>
       <concept_id>10010147.10010178.10010179.10010182</concept_id>
       <concept_desc>Computing methodologies~Natural language generation</concept_desc>
       <concept_significance>300</concept_significance>
       </concept>
    <concept>
       <concept_id>10003120.10011738.10011775</concept_id>
       <concept_desc>Human-centered computing~Accessibility technologies</concept_desc>
       <concept_significance>300</concept_significance>
       </concept>
 </ccs2012>
\end{CCSXML}

\ccsdesc[500]{Computing methodologies~Activity recognition and understanding}
\ccsdesc[300]{Computing methodologies~Natural language generation}
\ccsdesc[300]{Human-centered computing~Accessibility technologies}
 
\keywords{Sign Language Translation, Large Language Models}

\maketitle

\section{Introduction} 
\label{sec:introduction}
Current sign language (SL) understanding tasks mainly include  Continuous SL Recognition (CSLR)~\cite{gan2024signgraph} and SL Translation (SLT). CSLR aims to recognize a sequence of signs and convert it into the corresponding gloss sequence, while SLT~\cite{sincan2023context,gan2023towards} focuses on translating sign sequences into spoken language. As SLT produces more natural and fluent language outputs, it has attracted increasing attention in recent research. The de facto architecture for current SLT tasks typically employs a 2D or 3D CNN-based backbone to extract visual features~\cite{gan23contrastive}, followed by temporal modeling modules (\eg 1D CNNs or LSTMs) to capture dynamic changes across sign frames. Finally, a translation model (e.g., mBART~\cite{gan2025mixsigngraph,gan2026sign, ganlearning}, GPT-2~\cite{wong2024sign2gpt}) is used to generate the corresponding spoken language sentence.  SLT research has branched into two distinct paths: gloss-based SLT (GBSLT) and gloss-free SLT (GFSLT). 
Previous state-of-the-art (SOTA) GBSLT models~\cite{chen2022two, gan23contrastive} have  emphasized that pretraining the visual encoder with gloss labels is critical for achieving strong SLT performance.  Due to the reliance on gloss labels, SLT has increasingly shifted its focus towards GFSLT~\cite{wong2024sign2gpt, liang2024llava, gan2025mixsigngraph}, which aims to boost SLT performance without gloss supervision.

Meanwhile Large Language Models (LLMs) have revolutionized the field of natural language processing, setting new benchmarks across a broad spectrum of tasks~\cite{ataallah2024minigpt4, Maaz2023VideoChatGPT, gaskell2024mbot,guevarra2025llm}. Benefiting from the vast pretraining datasets, LLMs demonstrate extraordinary generalization capabilities, and as a result, fine-tuning  LLMs for downstream tasks has emerged as a prevailing trend~\cite{Maaz2023VideoChatGPT,qian2024momentor}.  However, adapting LLMs for the GFSLT task remains understudied~\cite{wong2024sign2gpt, gong2024llms}, primarily due to the following two key challenges.   
\textbf{1)} There exists a fundamental distributional gap between textual features and SL video features. Textual inputs, which are the primary input modality during LLM pretraining, are typically tokenized into discrete, symbolic units, forming clearly segmented sequences.  In contrast, SL video features are inherently continuous, with subtle variations between consecutive frames.  This discrepancy in representation poses a significant challenge in effectively tokenizing SL videos and providing suitable inputs for LLMs, which is crucial for fine-tuning LLMs for GFSLT.
\textbf{2)} Existing Vision-Language Models (\eg MiniGPT-4~\cite{ataallah2024minigpt4}) and LLM-based SLT models~\cite{liang2024llava} typically adopt a straightforward strategy by concatenating visual and textual features directly in an autoregressive manner. However, this approach often leads to the deprioritization of visual features and biases the model toward fitting the distribution of text tokens rather than visual ones~\cite{xiao2024seeing}, as LLMs are mainly pretrained on text-centric data, and visual inputs constitute an out-of-distribution modality.

To tackle the first challenge, a common approach in vision-language models (VLMs) is to align visual and textual features by employing a pre-trained image encoder and leveraging large-scale image-text datasets with objectives such as Image-Text Contrastive (ITC) loss and Image-Text Matching (ITM) loss~\cite{Maaz2024VideoGPT+,ye2024improving}. In addition, some works have explored using attention mechanisms~\cite{yin2023gloss}, vector quantization~\cite{gong2024llms}, and pseudo glosses~\cite{gan2025mixsigngraph, wong2024sign2gpt} to tokenize SL videos.  As for the second challenge, the effectiveness of concatenating visual and textual features in an autoregressive way remains underexplored.  CAL~\cite{xiao2024seeing} addresses this by focusing on reweighting text tokens based on their visual relevance, thereby enhancing the alignment between modalities. However, it overlooks the enhancement of visual representations themselves. In contrast, our approach aims to prioritize the contribution of visual features during prediction, ensuring that the model more effectively leverages visual information.

In this paper, we focus on the Gloss-Free SLT task by adopting open-source LLMs (e.g., the Llama family~\cite{touvron2023llama}). First, inspired by recent works~\cite{gan2025mixsigngraph, gong2024llms} and GBSLT  approaches where CTC loss is commonly used to pretrain the visual encoder, we propose a simple yet effective method to bridge the distributional gap between video and textual features, namely filtered pseudo-gloss CTC Pretraining. Specifically, we generate filtered pseudo-gloss sequences from the target text by applying lemmatization, removing prepositions and conjunctions, and randomly dropping tokens. The resulting pseudo-gloss sequences are then used with CTC loss to pretrain the visual encoder. 
To further prioritize the model's reliance on visual features, we propose Vision-Prioritized Distillation (VPD). This training strategy distills knowledge from a `Visual-Textual Prediction' path into a `Visual-Only Prediction' path, enabling the LLMs to generate text with greater reliance on visual features. By prioritizing visual features, VPD can boost the ability of text generation from visual inputs, thereby improving the performance of GFSLT models. We make the following contributions:

\begin{itemize} 
    \item We introduce a simple yet effective pretraining method, \textbf{Filtered  Pseudo-gloss CTC Pretraining (FPG-CTC)}, which generates filtered pseudo-gloss sequences to supervise the visual encoder. We show that FPG-CTC can effectively discretize and align SL features, leading to substantial performance improvements on the GFSLT task.

    \item We propose a \textbf{Vision-Prioritized Distillation (VPD)} training method, which distills the `Visual-Textual Prediction' into the `Visual-Only Prediction' to enhance GFSLT performance. The proposed VPD can boost text generation and alleviate the exposure bias by prioritizing visual features.
      
    \item We contribute an array of pre-trained SignLlama models to the community, ranging from 1 billion to 13 billion parameters. Extensive experiments on public SL datasets demonstrate the effectiveness of our models, which achieve very competitive performance on the GFSLT task.

\end{itemize}

\section{Related Work}
\label{sec:related}
\paragraph{LLMs for Visual  Understanding.} Open-source LLMs have revolutionized a wide range of downstream tasks, such as image understanding~\cite{li2023blip,wang2024cogvlm,chen2023minigptv2,liu2023visual} and video understanding~\cite{yang2023vid2seq}. These methods typically use a pretrained visual backbone (e.g., ViT~\cite{dosovitskiy2020vit}) to extract image or video features, which are then concatenated with textual inputs and fed into LLMs for text generation.  The training of these models typically follows a two-stage paradigm~\cite{li2023videochat}. First, they are tuned on large-scale video-text data to align the visual features with textual features. The second stage involves instruction tuning with data such as videos with descriptions and question-answering pairs. Such a paradigm allows the visual backbone to effectively tokenize images or videos and provide suitable inputs for LLMs. Although these methods have achieved promising results, transferring these pretrained backbones directly to SL tasks remains challenging. Unlike visual understanding tasks that often focus on scene-level changes, SL tasks involve much finer-grained visual cues, such as subtle hand movements, hand shapes, and facial expressions, which are often overlooked in general visual understanding. Moreover, a key limitation lies in the common practice of concatenating video and text features for autoregressive fine-tuning, which often leads the model to over-rely on textual features, deprioritize visual information, and ultimately degrade model performance.

\begin{figure*}[t]
	\centering           
 \includegraphics[width=0.90\textwidth]{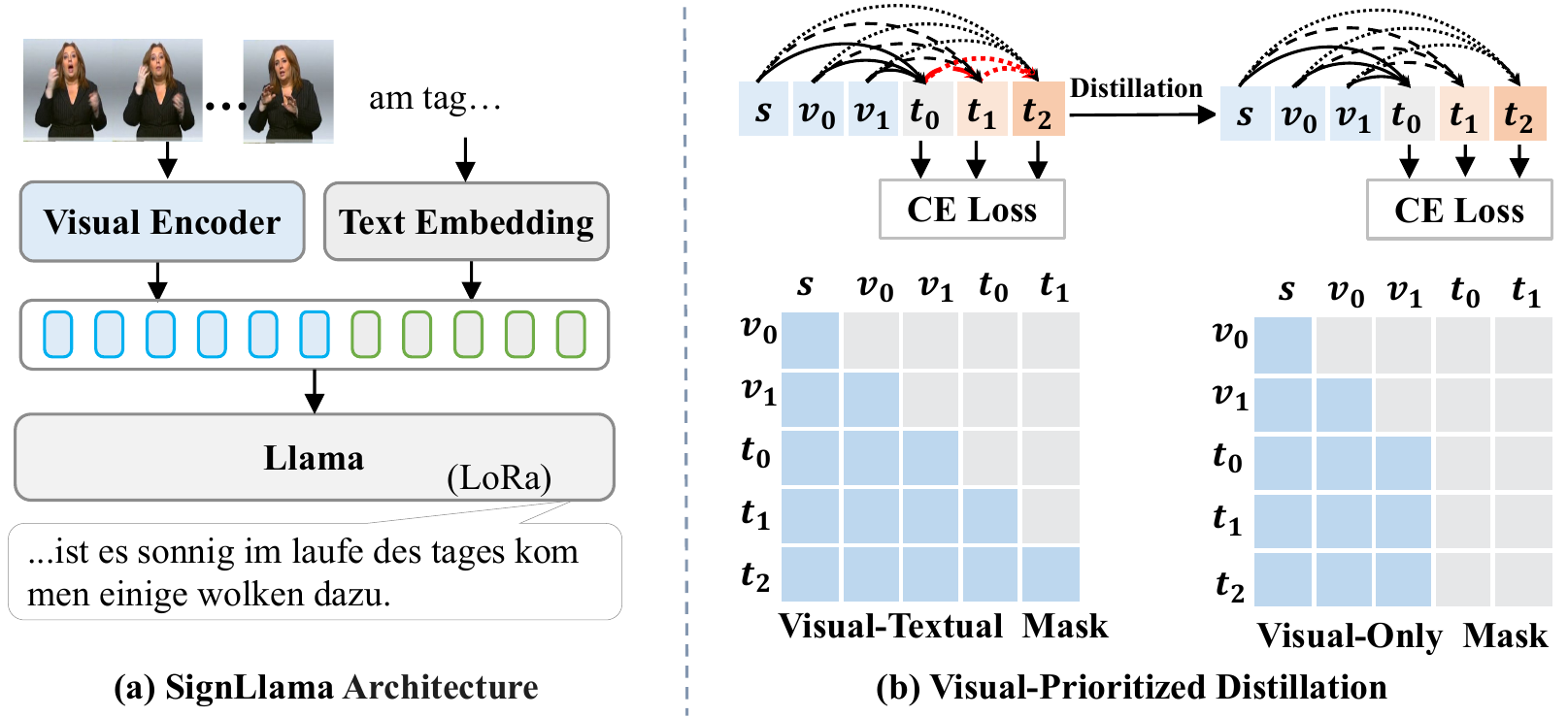} 
        \caption{The proposed SignLlama with  Visual-Prioritized Distillation} 
	\label{fig:SignLlama}  
\end{figure*}

\paragraph{Sign Language Tasks.}
Current approaches in CSLR~\cite{zuo2022c2slr,gan2024signgraph, chen2025c} and SLT~\cite{gan2021skeleton,zhou2021improving,uthus2023youtube, liu2026signpr} have established a de facto architecture, which typically involves employing a visual module to capture visual features, followed by a temporal module (\eg 1D CNNs, LSTM, or transformer layers) to model temporal dependencies. For CSLR tasks, a CTC loss is applied to compute the probabilities over all possible alignment paths. For SLT tasks, an additional translation model is employed to generate textual sequences from video features. Consequently, current CSLR research primarily focuses on improving the modeling of SL-related visual features and temporal relationships by designing various visual backbones (\eg VAC~\cite{min2021visual}, TwoStream~\cite{chen2022two}, SignGraph~\cite{gan2024signgraph}and MixSignGraph~\cite{gan2025mixsigngraph}) and temporal modules~\cite{zhu2024multiscale, zhang2014threshold}. Benefiting from CSLR, gloss-based SLT~\cite{gan23contrastive,gan2021skeleton} leverages pretrained backbones from the CSLR task and incorporates a translation model~\cite{liu2020multilingual, radford2019language} to generate text sequences. These pretrained backbones effectively align SL video features with gloss sequences, functioning as a tokenizer for SL videos. However, in GFSLT, a key challenge lies in how to effectively tokenize SL videos and align visual features with textual features to provide suitable inputs for LLMs, which is crucial for fine-tuning LLMs for SLT. Existing GFSLT models typically adopt attention mechanisms~\cite{yin2023gloss} or contrastive language-image pretraining~\cite{jiao2024visual, liang2024llava} to enhance alignment and improve performance. Recently, SignLLM~\cite{gong2024llms} attempted to address this by training two discrete codebooks to align sign video features with text. There are also some works focusing on designing pseudo gloss~\cite{gan2025mixsigngraph, ganlearning} to pretrain their backbone, while our methods provides a simpler, more straightforward, and more effective way to pretrain SL backbones. 
Nonetheless, these methods primarily focus on how to effectively tokenize SL video inputs.  In addition, recent GFSLT models incorporating LLMs also typically adopt the same strategy, \ie, concatenating video and textual features as input for autoregressive fine-tuning. However, this approach tends to prioritize textual features over visual ones.  In this paper, we aim to develop a simple and effective training paradigm for effectively tokenizing SL video, along with a strategy that enhances the GFSLT model performance by prioritizing visual features.

\section{Method}
\label{sec:Method}
\paragraph{Overall Framework.}   
For an SL video with $\theta$ frames $f=\{f_i\}_{i=1}^{\theta}$, the goal of GFSLT is to generate  $t=\{t_i\}_{i=1}^ { \varsigma }$ with $\varsigma$ words based on input $f$. As illustrated in Figure~\ref{fig:SignLlama}, our model consists of  a visual encoder and a Llama model. Note that the text embedding module is part of Llama. We show it explicitly here for clarity. The visual encoder $\mathcal{VE}$ processes frames to get visual features ${v}$=$\mathcal{VE}({f})$ with $m$ vectors, where ${v}=\{v_{i}\in \mathbb{R}^{d_I}\}_{i=1}^{m}$ ($m\leq \theta $) and $d_I$ is the input dimension of Llama. The tokenizer converts the target text into $n$ token IDs $\{u_i\}_{i=1}^{n}$, and the text embedding module maps them to
text embeddings $e=\{e_i\in\mathbb{R}^{d_I}\}_{i=1}^{n}$. ($n \ge \varsigma$). During training stage, the model predicts  token $\hat{u}_i$ based on $v$  and previous ground truth tokens: $p(\hat{u}_i| v, \{u_j\}_{j=0}^{i-1})$. During inference, the model predicts  token $\hat{t}_i$ based on $v$  and previous predicted tokens: $p(\hat{u}_i| v, \{\hat{u}_j\}_{j=0}^{i-1})$.

\subsection{Filtered Pseudo-gloss CTC Pretraining}
For GBSLT~\cite{chen2022two, gan2025mixsigngraph}, the visual encoder is typically trained with the CSLR task, in which CTC loss is used to optimize the visual encoder to learn semantic segmentation and alignment with gloss sequences $g$. A translation model is then integrated and fine-tuned for SLT, taking the output of the  encoder as its input. Let 
$R$ denote the recognition network (with the visual encoder as a subcomponent) and $T$ denote the translation network, while $\Theta_R$ and $\Theta_T$ denote the parameters of the recognition and translation networks, respectively. $\mathcal{L}_{CTC}$ and $\mathcal{L}_{CE}$ denote the CTC loss and the cross-entropy loss.
%CTC loss is denoted as $\mathcal{L}_{CTC}$, and the cross-entropy loss is denoted by $\mathcal{L}_{CE}$. 
The hyperparameters $\alpha$ and $\beta$ are used to balance the two objectives. The training process can be formalized as:

\begin{gather}  
 \label{equ:ctcloss} 
 \min_{\Theta_R} \mathcal{L}_{CTC}(R(f), g)  \\
 \label{equ:CEloss} 
 \min_{\Theta_R, \Theta_T} (\alpha \mathcal{L}_{CTC}(R(f), g) + \beta \mathcal{L}_{CE}(T(v), t)) 
\end{gather}

For gloss-free SLT, the absence of gloss annotations makes it challenging to pretrain the visual encoder. Unlike previous pseudo gloss based GFSLT models that adopt contrastive learning for visual backbone pretraining, we aim to develop a simple yet effective method that enables the visual encoder to effectively tokenize SL videos and align visual features with textual representations. We draw inspiration from previous work~\cite{gan2025mixsigngraph} and propose the Filtered  Pseudo-gloss CTC Pretraining (FPG-CTC) method. Specifically,  we generate a pseudo-gloss sequence 
$g_p$ from the target text by applying lemmatization and filtering out prepositions and conjunctions.  Considering that certain words in the text may not be explicitly expressed in the corresponding SL video, we further apply random dropping to remove a fixed proportion of words, resulting in the final pseudo-gloss sequence $g_p$. The FPG-CTC training process can be formalized as:

\begin{gather}   
 \label{equ:ctcloss2} 
 \min_{\Theta_R} \mathcal{L}_{CTC}(R(f), g_p)  \\
 \label{equ:CEloss2} 
 \min_{\Theta_R, \Theta_T} (\alpha \mathcal{L}_{CTC}(R(f), g_p) + \beta \mathcal{L}_{CE}(T(v), t))  
\end{gather}

\subsection{Visual-Prioritized Distillation Training for LLM Fine-Tuning}
After obtaining visual features, a widely adopted approach is to concatenate visual and textual features directly, followed by next-token prediction, which we refer to as `Visual-Textual Prediction', formalized as: $p(\hat{t}_i|v, \{t_j\}_{j=0}^{i-1})$.  However, this training paradigm suffers from a critical limitation: visual deprioritization. LLMs tend to prioritize fitting the distribution of text tokens over that of the newly introduced visual tokens. This is because LLMs are pretrained on text-centric data, and visual inputs constitute an out-of-distribution modality, leading to the deprioritization of visual information during generation.  During inference, the model no longer has access to ground-truth tokens $\{t_j\}_{j=0}^{i-1}$ and instead conditions on its own predictions: $p(\hat{t}_i| v, \{\hat{t}_j\}_{j=0}^{i-1})$. This mismatch between training and testing leads to cumulative errors. When the model is overly reliant on generated text, errors in early predictions can quickly accumulate, leading to degraded performance.
  
To encourage the LLMs to prioritize visual features, we propose a novel training paradigm termed Visual-Prioritized Distillation. Specifically, we perform two separate forward passes through the LLM: one for Visual-Textual Prediction, where visual and textual features are concatenated as input; and another for Visual-Only Prediction, where the textual inputs are masked out using a `Visual-Only Mask' (as illustrated in Figure~\ref{fig:SignLlama} (b)), and the model is required to generate the full target sequence relying solely on visual inputs. To ensure that the `Visual-Only  Prediction' path can converge and perform well, we distill the logits from the `Visual-Textual  Prediction' path into the `Visual-Only Prediction' one. This distillation guides the learning process of the visual-only prediction path, encouraging it to align with the output distribution of the visual-textual prediction  path  while learning to rely exclusively on visual features. The training loss can be formalized as:  
 
\begin{equation} 
\begin{aligned}
    \mathcal{L}_{VPD} &=  \mathcal{L}_T^{vt} +  \mathcal{L}_T^{v} +  \gamma \mathcal{L}_{KD} \\
  &=\sum_{i=1}^{n}\mathcal{L}_{CE}(T(v, \{t_j\}_{j=0}^{i-1}),  t_i) \\
  &+ \sum_{i=1}^{n}\mathcal{L}_{CE}(T(v),  t_i)  + \lambda  \mathrm{KL}(T(v, t)||T(v))  
\end{aligned}
 \label{equ:VPD} 
\end{equation}
where $\gamma$ is a hyperparameter and $n$ denotes the number of tokens. $ \mathcal{L}_T^{vt} $ , $ \mathcal{L}_T^{v} $ are  cross-entropy losses of `Visual-Textual Prediction' and `Visual-Only  Prediction',  and  $\mathcal{L}_{KD}(T(v, t), T(v))$ 
 denotes the distillation loss. 
  
\paragraph{Training and Inference for GFSLT}
The final training process of our model begins with pretraining the visual backbone using pseudo-gloss supervision, as described below.
\begin{equation}   
 \mathcal{L}_{R} = \min_{\Theta_R} \mathcal{L}_{CTC}(R(f), g_p) 
 \label{equ:pgctc} 
\end{equation} 
The entire model is then fine-tuned using the VPD objective together with the FPG-CTC loss.
\begin{equation}  
  \min_{\Theta_R, \Theta_T} (\alpha \mathcal{L}_{R} +  \beta (\mathcal{L}_T^{vt} +  \mathcal{L}_T^{v} +  \lambda \mathcal{L}_{KD} )) 
 \label{equ:finallloss} 
\end{equation}  
During inference, the visual encoder takes the SL video as input and extracts SL video features, which are then fed into the Llama model. The Llama model follows the standard autoregressive process to generate the target text sequence. Neither pseudo-gloss supervision nor the VPD training procedure is required during inference.

\section{Theoretical Insights into VPD}
\paragraph{The difference between VPD and Knowledge Distillation} 
Although VPD shares a superficial similarity with knowledge distillation (KD), it is conceptually distinct in both purpose and design. The key novelty lies in how VPD prioritizes visual features between visual-only and visual-textual paths, mitigating the tendency of the AR model to over-rely on textual cues and underutilize visual information, which is a critical bottleneck that has not been addressed in prior work.  Thus, VPD is not merely an adaptation of KD, but a novel mechanism that explicitly prioritizes visual features in SLT.

\paragraph{The theoretical intuition of VPD} Here, we also provide a brief theoretical intuition of VPD at the gradient level. Let $v$ denote the visual input encoded by parameters $\theta_v$, $t$ the optional textual input, $T(v,t)$ the teacher (visual-textual path) with logits $z_{vt}$ and softmax probabilities $p_{vt}$, and $S(v)$ the student (visual-only path) with logits $z_v$ and softmax probabilities $p_v$. The simplified VPD loss is:

\begin{align}
   \mathcal{L} = \underbrace{\mathcal{L}_{vt}}_{\text{visual-textual CE}} + \underbrace{\mathcal{L}_{v}}_{\text{visual-only CE}} + \gamma\underbrace{\mathcal{L}_{KD}}_{\text{distillation (KL)}}, 
\end{align} 
where
\begin{equation}
   \mathcal{L}_{vt} = -\sum_{i} y_i \log p_{vt,i},
\end{equation}
\begin{equation}
   \mathcal{L}_{v}  = -\sum_{i} y_i \log p_{v,i},
\end{equation}
\begin{equation}
   \mathcal{L}_{KD} = \mathrm{KL}(p_{vt}\,\|\,p_{v})
= \sum_{i} p_{vt,i}\log\frac{p_{vt,i}}{p_{v,i}}. 
\end{equation}
Where y denote the one-hot target distribution. We are interested in the gradient of $\mathcal{L}$ with respect to $\theta_v$:
\begin{align}
   \frac{\partial \mathcal{L}}{\partial \theta_v} 
= \underbrace{\frac{\partial \mathcal{L}_{vt}}{\partial \theta_v} + \frac{\partial \mathcal{L}_v}{\partial \theta_v}}_{\text{CE gradients}} + \gamma \sum_i (p_{v,i}-p_{vt,i}) \frac{\partial z_{v,i}}{\partial \theta_v}.
\end{align} 
Intuitively, when the teacher assigns a higher probability to the correct class, the KL term provides an additional corrective gradient that pushes
$p_v$ toward $p_{vt}$, supplementing the supervision provided by the
visual-only CE loss. This provides an additional training signal to the visual encoder, encouraging
the visual-only path to produce predictions closer to those of the
visual-textual path.

\section{Experiments}
\label{sec:Experiments}

\paragraph{Datasets} 
We evaluate our model on several publicly available SLT datasets, including the widely used Phoenix14T and CSL-Daily, as well as two large-scale ASL datasets: How2Sign and OpenASL. (1) \textbf{Phoenix14T} \cite{cihan2018neural}  is a German Sign Language dataset annotated with both glosses and translations. It consists of 7096 training, 519 validation, and 642 test samples from 9 signers, with a vocabulary of 1066 glosses and 2877 German words.  (2) \textbf{CSL-Daily} \cite{zhou2021improving}  is a Chinese SL dataset containing 18401, 1077, and 1176 videos for training, validation, and testing from 10 signers. It provides 2000 glosses and 2343 words for translation.  (3) \textbf{How2Sign}  \cite{duarte2021how2sign} is a large-scale American Sign Language (ASL) dataset comprising over 80 hours of multi-view video and multimodal data. We use only the frontal-view RGB videos, including 31128, 1741, and 2322 samples for training, validation, and testing, respectively.  (4) \textbf{OpenASL} \cite{shi2022open} is another large ASL dataset with over 280 hours of video from more than 200 signers. It includes 96476 training, 997 validation, and 999 test samples.

\paragraph{Data Preprocessing}
\label{sec:experimental_setting}
Following prior work~\cite{min2021visual, gan2024signgraph, gan2025mixsigngraph}, we adopt the same preprocessing pipeline to ensure fair comparisons. During training, we apply standard data augmentations, including resizing frames to 256$\times$256 pixels, random cropping to 224$\times$224 pixels, random horizontal flipping with a probability of 0.5, and random temporal scaling within ±20\%. During inference, frames are resized to 256$\times$256 and center-cropped to 224$\times$224.

\begin{table}[t] 
\caption{Effects of our training strategy in Phoenix14T.}
\tablestyle{1.5pt}{1.2} 
	\centering 
	\resizebox{0.95\columnwidth}{!}{ 
		\begin{tabular}{cc|ccccc}
		\shline 
\multirow{2}*{FPG-CTC}&\multirow{2}*{VPD}& \multicolumn{5}{c}{TEST} \\ 
&&ROUGE& BLEU1& BLEU2& BLEU3& BLEU4 \\
			\shline 
 &  &   20.09 &18.12&8.56&6.46&5.46 \\ 
\checkmark & & 49.93 &49.22&36.53 &28.61 &23.47 \\ 
\checkmark  &\checkmark & \underline{52.81} &\underline{52.54}&\underline{40.96} &\underline{32.87}&\underline{26.74}\\	
	\shline 
	\end{tabular}} 
	\label{tab:fgvpd}  
\end{table}
  
 \begin{table}[t]
	\centering 
\tablestyle{1.5pt}{1.2} 
	\caption{Effects of VPD loss in Phoenix14T.}
	\resizebox{0.93\columnwidth}{!}{ 
		\begin{tabular}{ccc|cccccc}
		\shline  
\multirow{2}*{$\mathcal{L}_{T}^{vt}$}&\multirow{2}*{$\mathcal{L}_{T}^{v}$}&\multirow{2}*{$\mathcal{L}_{KD}$}& \multicolumn{5}{c}{TEST} \\
&&&ROUGE& BLEU1& BLEU2& BLEU3& BLEU4 \\
			\shline  
\checkmark& & 
&49.93 &49.22&36.53 &28.61 &23.47\\	

&\checkmark & 
&49.72  &49.75 &36.51 &28.30 &23.25\\	

\checkmark&\checkmark &
&51.08& 50.16&38.08&29.51&23.82\\	

\checkmark  &\checkmark  &   \checkmark 
&\underline{52.81} &\underline{52.54}&\underline{40.96} &\underline{32.87}&\underline{26.74}\\	
\shline   
\end{tabular}}
\label{tab:proposed_loss}   
\end{table}

\paragraph{Implementation Details} Our architecture consists of following key components: (1) \textit{Visual Encoder}: In our baseline setting, we choose PoolFormer~\cite{yu2022metaformer} with 1D CNN layers used in previous  work~\cite{gan2024signgraph, min2021visual}. (2) \textit{Llama}: We adopt the Llama-3.2-1B model\footnote{\url{https://huggingface.co/meta-llama/Llama-3.2-1B}} in our baseline setting. We choose the base pretrained model rather than instruction-tuned variants (\eg Llama-3.1-1B-Instruct), as the effects of prompt engineering are beyond the scope of this work. Instead, our focus is on improving the performance of base LLMs in the context of gloss-free SLT.  For LoRA used in Llama, we apply LoRA~\cite{hu2022lora} with a rank of 4 and a scaling factor (`LoRA alpha') of 32. LoRA adapters are inserted into the  `q-proj', `v-proj', and `o-proj'  layers. In addition, we fully fine-tune both the input embedding layer and the output layer in Llama, which are included in the list of trainable modules.  (3) \textit{Training Setting}: We train the model using the Adam optimizer with a weight decay of 0.0001 for 50 epochs on three GeForce RTX 3090 GPUs. The initial learning rate is set to 1e-4 for the recognition model and 1e-5 for the translation model, and is decayed by a factor of 0.5 at epochs 10, 20 and 30. The batch size is set to 3. For FPG-CTC pretraining, the random dropout rate is set to 0.2. The loss weights   $\alpha$, $\beta$ in Equation~\ref{equ:finallloss} are set to 1,  and $\lambda$ is set to 5. To ensure the model can be trained end-to-end within 24GB of GPU memory, the entire model is trained in half-precision.

\paragraph{Evaluation Metrics}
To evaluate our model, we adopt  the ROUGE-L F1 Score~\cite{lin2004rouge}, BLEU-1,2,3,4~\cite{papineni2002bleu} and BLEURT~\cite{sellam2020bleurt} for gloss-free SLT, which are common metrics in existing work~\cite{gan2021skeleton, chen2022two}.

\begin{table}[t]
	\centering 
\caption{Effect of pseudo-gloss. POS means Part-Of-Speech.} 
	\resizebox{0.89\linewidth}{!}{
\begin{tabular}{l|ccc} 
\hline
{Model}&{ROUGE}&BLEU1 &BLEU4  \\
\toprule 
w/o lemmatizer        &50.19 & 51.62& 24.74  \\
 w/o stopword   & 50.00& 50.04& 24.16\\
 POS-only & 33.00& 31.41&18.34 \\
Ours   & \underline{ 52.81 }&\underline{52.54} & \underline{26.74}\\
\hline
\end{tabular}}
\label{tab:pusdo-gloss}
\end{table}

 \begin{figure}[t]  
\centering 
\includegraphics[width=0.48\linewidth]{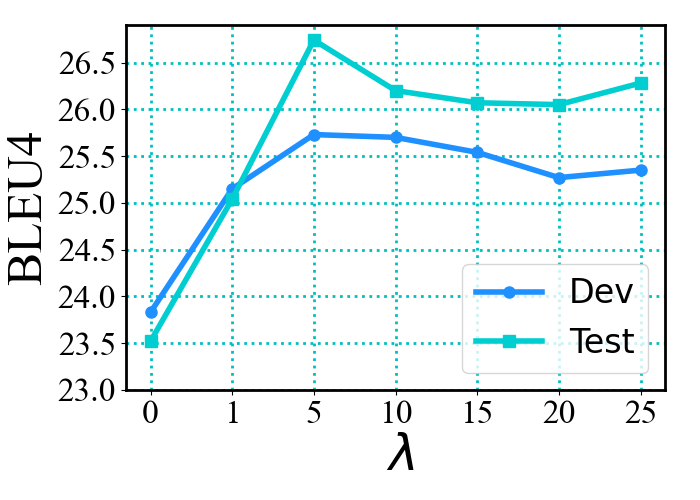}
\includegraphics[width=0.48\linewidth]{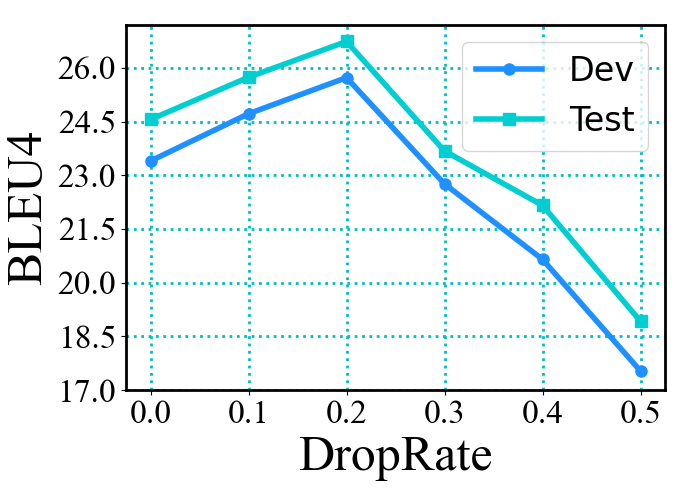}  
\caption{Effect of different $\lambda$ in VPD and different drop rates in FPG-CTC} 
\vspace{-2mm}
\label{fig:droprate} 
\end{figure} 

\section{Ablation Study}\label{sec:ablation}
Following previous work~\cite{cihan2018neural, chen2022two, gan2024signgraph}, we perform ablation studies on the Phoenix14T dataset to verify the effectiveness of  our proposed SignLlama model.

\paragraph{Effect of the Proposed Training Strategies.}
 To validate the effectiveness of our proposed FPG-CTC pretraining and VPD training strategies, we compare our approach against baselines that are trained via direct fine-tuning. The results are shown in Table~\ref{tab:fgvpd}.  The direct fine-tuning baseline 
 model  performs poorly, achieving only 5.46 BLEU-4 on the `Test' sets. In contrast, applying FPG-CTC  significantly boosts performance, demonstrating that our pseudo-gloss CTC pretraining effectively trains the visual encoder and facilitates better alignment between visual and textual features. 
Combining FPG-CTC with VPD yields the best results, which suggests that prioritizing visual features leads to better performance.

\paragraph{Effect of VPD Loss.} The proposed VPD loss comprises three components: $\mathcal{L}_{T}^{vt}$, $\mathcal{L}_{T}^{v}$, and $\mathcal{L}_{KD}$. As shown in Table~\ref{tab:proposed_loss}, training with $\mathcal{L}_{T}^{vt}$ alone (\ie standard autoregressive learning with both visual and textual inputs) achieves BLEU-4 scores of 23.47 on the test set. In comparison, using only $\mathcal{L}_{T}^{v}$, which relies solely on visual inputs, yields slightly lower BLEU-4 score of 23.25. When the distillation loss $\mathcal{L}_{KD}$ is introduced, the model achieves substantial performance gains, reaching 26.74 BLEU-4 on the test set. These results highlight the effectiveness of VPD in enhancing visual feature utilization via KD, thereby improving overall translation quality.

\paragraph{Effect of Pseudo Gloss}
To evaluate the effect of different preprocessing steps on FPG-CTC, we conduct ablation studies, as shown in Table~\ref{tab:pusdo-gloss}. Removing either the lemmatization or stopword filtering results in a small  drop in performance. Using only POS information significantly degrades performance, showing that part-of-speech information alone is insufficient. These results indicate that our approach benefits from the combined preprocessing steps of lemmatization and stopword removal, while the overall trends remain consistent across different settings.

\paragraph{Effect of the Hyperparameter $\lambda$ in VPD}  
We introduce a hyperparameter $\lambda$ in the VPD loss to control the contribution of the distillation term. To analyze its impact, we conduct experiments with varying values of $\lambda$. As illustrated in Figure~\ref{fig:droprate}, the best BLEU-4 performance is observed when $\lambda$ is set to 5. Based on this observation, we set $\lambda=5$ in all experiments for better performance.

\begin{table}[t]
	\centering 
	\caption{Effect of different Llama pretrained model.}
	\resizebox{0.87\columnwidth}{!}{ 
		\begin{tabular}{l|ccc}
		\shline
\multirow{2}*{Model}&  \multicolumn{3}{c}{TEST} \\
& ROUGE& BLEU1& BLEU4 \\
			\shline 
Llama2 7B &{52.61} &{52.14}&{26.26} \\
Llama2 13B &{52.73} &{52.54}&{26.54} \\	
Llama3.2 1B &{52.81}  &{52.54} &{26.74}\\
Llama3.2 3B& 52.15 &52.61&26.97 \\	
Llama3.1 8B& 52.98 &52.55 &26.85 \\	
Llama3 8B& 52.40 &52.24 &26.82 \\	 
	\shline 
	\end{tabular} }   
	\label{tab:Llamaweight}   
\end{table}

\paragraph{Effect of Drop Rate in FPG-CTC} 
We apply random dropping during pseudo-gloss generation in FPG-CTC and further investigate the impact of varying dropout rates. As shown in Figure~\ref{fig:droprate}, model performance improves with moderate dropout rates but degrades when the dropout rate is too high, as excessive word removal hinders effective alignment learning between SL videos and pseudo-gloss sequences during FPG-CTC pretraining. Accordingly, we use a drop rate of 0.2 in all experiments.

\paragraph{Effect of Different Llama Sizes} 
Llama is a family of LLMs ranging from 1B to 405B parameters. Given our limited GPU resources, we scale our model from the 1B model to larger variants up to 13B, including both the Llama 2 and Llama 3 series. As shown in Table~\ref{tab:Llamaweight}, interestingly, increasing the size of the pretrained model yields only marginal improvements. This may be attributed to the fact that emergent behaviors often associated with LLMs tend to manifest at larger model scales.  \textbf{This conclusion is consistent with the findings in~\cite{zhang2024scaling}.} Since scaling the Llama model size does not yield significant improvements, we report results using the Llama 3.2 1B model in all subsequent comparisons, unless otherwise specified.

\paragraph{Training and Inference Speed.}
To evaluate VPD's impact on training time, we conduct experiments on three GeForce RTX 3090 GPUs with a batch size of 3, and limit the maximum training sample to 280 frames on Phoenix14T. As shown in Table~\ref{tab:trainingtime}, VPD introduces only a slight overhead, increasing the training time per epoch from 20 minutes to 22 minutes. This demonstrates that VPD brings negligible additional cost while providing performance gains.
Here, we also report the inference latency on a single NVIDIA RTX 3090 GPU with  approximately 250 frames for one video. 
As shown  in Table~\ref{tab:inference_time}, even the 13B model maintains reasonable inference efficiency, while the 1B version achieves 0.85s/video with one 3090 GPU.
   
\begin{figure}[t]
\centering 
\includegraphics[width=0.97\linewidth]{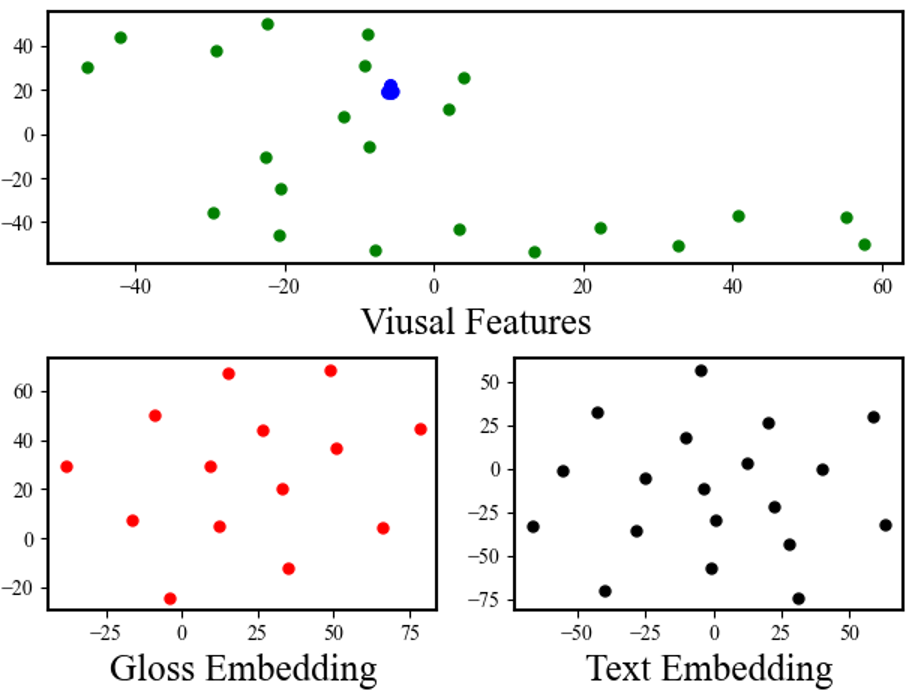} 
\caption{Visualization of feature distributions using t-SNE~\cite{van2008visualizing}. Blue and green points represent the visual features without/with FPG-CTC. Red and black points represent the corresponding gloss and text sequence embeddings of Llama.} 
\label{fig:tcp_vis}   
\end{figure}

\begin{table}[t]
\centering
\begin{minipage}{0.40\columnwidth}
\centering
\caption{Training speed.}
\resizebox{1\columnwidth}{!}{
\begin{tabular}{l|c} 
{Model}&{Training Time}  \\
\toprule 
w VPD        &22min/epoch  \\
 w/o VPD &20 min/epoch
\end{tabular}}
\label{tab:trainingtime} 
\end{minipage}
\hfill
\begin{minipage}{0.56\columnwidth}
\centering
\caption{Model inference speed.} 
\resizebox{1\columnwidth}{!}{ 
\begin{tabular}{l|rrrr} 
\multirow{2}*{Inference(s)}&\multicolumn{4}{|c}{LLama Model Size} \\  
&{1B}&{3B} &8B&13B \\ 
\toprule 
 250 frames &  0.85 &2.14&2.55&4.23    \\  
	\end{tabular}}   
\label{tab:inference_time}
\end{minipage}  
\end{table}

\begin{table*}[t]
% \tablestyle{3.0pt}{1.1}  
	\centering 
    	\caption{Comparison of SLT performance on Phoenix14T datasets.}
	\resizebox{0.95\textwidth}{!}{
\begin{tabular}{l|ccccc|ccccc}  
\shline
\multirow{3}*{Gloss free SLT}& \multicolumn{10}{c}{Phoenix14T}   \\
&\multicolumn{5}{c|}{DEV} & \multicolumn{5}{c}{TEST}\\ 
&ROUGE &BLEU1& BLEU2& BLEU3& BLEU4& ROUGE& BLEU1& BLEU2& BLEU3& BLEU4 \\
\shline  
SignCL~\cite{ye2024improving}& & & & & &49.04&49.76&36.85&29.97&22.74\\
Sign2GPT\cite{wong2024sign2gpt}&  -&-&-&-&-&48.90&49.54&35.96&28.83&22.52\\
GFSLT-VLP-SignCL~\cite{ye2024improving}&  -&-&-&-&-&  49.04 &49.76& 36.85& 29.97& 22.74 \\
LLaVA-SLT~\cite{liang2024llava}&-&-&-&-&-&50.44& 51.20& 37.51& 29.39 &23.43\\ 
FLa-LLM~\cite{chen2024factorized} &-&-&-&-&-&45.27 &46.29& 35.33& 28.03& 23.09\\
C2RL~\cite{chen2025c}&-&-&-&-&- &50.96& 52.81& 40.20& 32.20& 26.75\\ 
SignLLM~\cite{gong2024llms}& {44.49}&{46.88}&{36.59}&{29.91}&{25.25}&{47.23}&{45.21}&{34.78}&{28.05}&{23.40} \\ 
MixSignGraph~\cite{gan2025mixsigngraph}& 51.71 &51.07& 37.97& 29.98& 24.87& 51.14& 50.01& 38.04& 29.95& 24.02 \\
{SignDINO}~\cite{ganlearning}&52.36&{53.64}&{38.65}&{{30.49}}&{{25.62}}&{{52.75}}&{{52.13}}&{{39.64}}&{{33.73}}&{{25.46}}\\  
\baseline{SignLlama 1B} &  \baseline{52.53} & \baseline{52.13}&\baseline{38.88}&\baseline{30.61} &\baseline{25.73} &\baseline{52.81} &\baseline{52.54}&\baseline{40.96} &\baseline{32.87}&\baseline{26.74}\\
\shline
	\end{tabular}}    
	\label{tab:SLTPhoenix14T}   
\end{table*}

\begin{table*}[t]
	\centering  
	\caption{Comparison of SLT performance on CSL-Daily datasets.}   
	\resizebox{0.95\textwidth}{!}{
\begin{tabular}{l|ccccc|ccccc} 
\shline
\multirow{3}*{Gloss-free SLT}& \multicolumn{10}{c}{CSL-Daily }   \\
%\cline{2-11}
&\multicolumn{5}{c|}{DEV} & \multicolumn{5}{c}{TEST}\\ 
&ROUGE &BLEU1& BLEU2& BLEU3& BLEU4& ROUGE& BLEU1& BLEU2& BLEU3& BLEU4 \\
\shline 
% SL-Luong~\cite{cihan2018neural}&34.28&34.22&19.72&12.24&7.96&34.54&34.16&19.57&11.84&7.56\\
GASLT~\cite{yin2023gloss}&-&-&-&-&&20.35&19.90&9.94&5.98&4.07\\
GFSLT~\cite{zhou2023gloss}&35.16&37.60&23.30&14.89&9.92&35.42&37.69&23.28&14.93&9.88\\
GFSLT-VLP~\cite{zhou2023gloss}&36.44&39.20&25.02&16.35&11.07&36.70&39.37&24.93&16.26&11.00\\
SignCL~\cite{ye2024improving}& & & & & &48.92&47.47&32.53&22.62&16.16\\
Sign2GPT\cite{wong2024sign2gpt}&  -&-&-&-&-&42.36&41.75& 28.73& 20.60 &15.40 \\ 
SignLLM~\cite{gong2024llms}&  39.18&42.45&26.88&17.90&12.23&39.91&39.55&28.13&20.07& 15.75 \\ 
GFSLT-VLP-SignCL~\cite{ye2024improving}&  -&-&-&-&-& 48.92  &47.47& 32.53& 22.62& 16.16 \\ 
FLa-LLM~\cite{chen2024factorized} &-&-&-&-&-&37.25&37.13&25.12&18.38&14.20\\
C2RL~\cite{chen2025c}&-&-&-&-&- &48.21 &49.32&36.28&27.54&21.61\\
LLaVA-SLT~\cite{liang2024llava}&-&-&-&-&-& 51.26&52.15& 36.24& 26.47& 20.42 \\ 
MixSignGraph~\cite{gan2025mixsigngraph}& 49.16 &49.98& 36.42 &26.89& 20.43 &49.93 &50.24 &36.91 &27.54 &20.78 \\
\baseline{SignLlama}&\baseline{51.62}&\baseline{52.12}&\baseline{41.23}&\baseline{33.23}&\baseline{24.83}&\baseline{51.43}&\baseline{52.38}&\baseline{39.78}&\baseline{33.91}&\baseline{24.76}\\
\shline
	\end{tabular}} 
	\label{tab:SLTCSL-Daily}  
\end{table*}

\section{Qualitative Results}

\paragraph{FPG-CTC Visualization.} We  visualize the visual feature distribution of one test sample from Phoenix14T, trained with and without FPG-CTC, alongside the corresponding gloss/text sequence embeddings from Llama.
As shown in Figure~\ref{fig:tcp_vis},  the gloss/text inputs are tokenized into discrete tokens. In contrast, the visual feature distribution trained without FPG-CTC exhibits a non-discriminative and continuous pattern. When trained with FPG-CTC, the visual features are better tokenized and exhibit a more discriminative distribution, which closely resembles that of the textual input. This demonstrates that FPG-CTC effectively tokenizes SL videos and provides suitable inputs for the LLMs.

\begin{figure}[t]
\centering 
\includegraphics[width=0.48\linewidth]{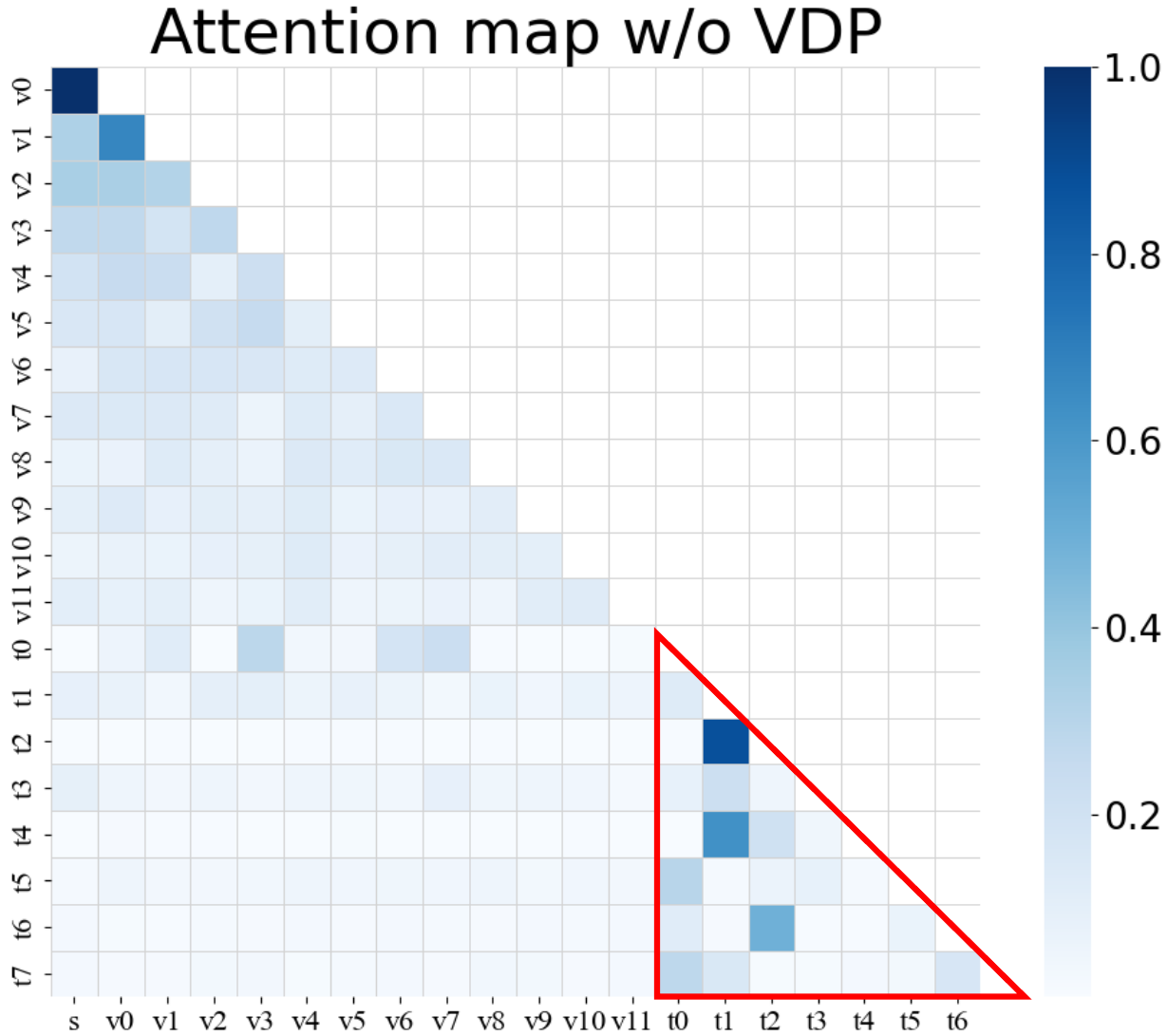} 
\includegraphics[width=0.48\linewidth]{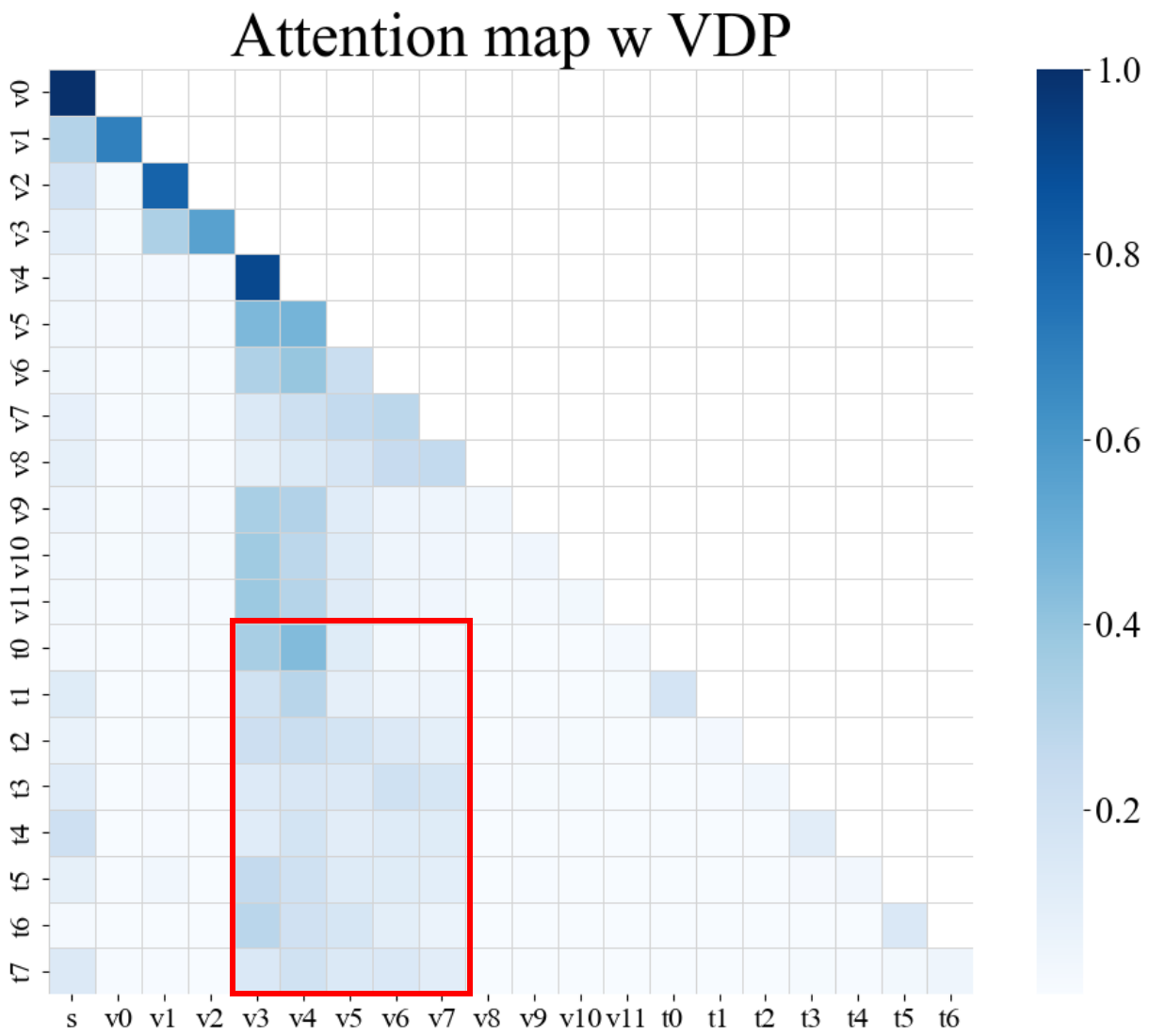} 
\caption{Visualization of the attention weights of Llama.}
\label{fig:VPD_Vis}   
\end{figure}

\paragraph{VPD Visualization.} To investigate the effect of VPD, we visualize one of the attention heads from the final layer of the Llama model (other heads and layers exhibit similar patterns) under models trained with and without VPD. As shown in Figure~\ref{fig:VPD_Vis}, ‘s’ denotes the start-of-sequence token, ‘v0–v11’ represent visual tokens, and ‘t0–t7’ indicate text tokens. The model trained without VPD primarily relies on previous text tokens for next-token prediction, with limited attention to visual inputs. In contrast, the model trained with VPD allocates significantly more attention to the visual tokens. This demonstrates that VPD effectively prioritizes visual features for LLMs, thereby improving translation performance.

\begin{table*}[h]
	\centering  
 \caption{Comparison of GFSLT performance on  How2Sign.} 
	\resizebox{0.95\textwidth}{!}{
\begin{tabular}{l|ccccc|ccccc} 
\shline
\multirow{3}*{Gloss-free SLT}& \multicolumn{10}{c}{How2Sign}   \\
%\cline{2-11}
&\multicolumn{5}{c|}{DEV} & \multicolumn{5}{c}{TEST}\\ 
&ROUGE &BLEU1& BLEU2& BLEU3& BLEU4& ROUGE& BLEU1& BLEU2& BLEU3& BLEU4 \\
\shline   
{YouTube-SLT~\cite{uthus2023youtube}}&{-}&{-}&{-}&{-}&{-}&-
&{14.96} &{5.11} &{2.26} &{1.22}\\ 
{YouTube-SLT-P~\cite{uthus2023youtube}}& {-} &{-}&{-}&{-}&{-}&{-}&{37.82} &{24.13} &{16.92} &{12.39}\\
{SSVP-SLT~\cite{rust2024towards}} &{-}&{-}&{-}&{-}&{-}&25.70
&{30.20} &{16.70} &{10.50} &{7.00}\\ 
$C^2$RL~\cite{chen2025c}&-&-&-&-&- &27.81 &29.81 &18.99 &13.27& 9.66\\ 
FLa-LLM~\cite{chen2024factorized}&-&-&-&-&-& 27.81 &29.81 &18.99 &13.27& 9.66\\
GloFE-VN~\cite{lin2023gloss}&12.98 &15.21 &7.38& 4.07& 2.37&12.61 &14.94& 7.27& 3.93& 2.24\\ 
SLT-IV~\cite{tarres2023sign}&-&35.20& 20.62& 13.25& 8.89&-&34.01 &19.3& 12.18&8.03\\
MixSignGraph~\cite{gan2025mixsigngraph}&25.41& 26.82 &16.70& 11.48& 8.36 &25.71& 26.65& 16.55& 11.68& 8.69\\ 
\baseline{SignLlama 1B}& \baseline{30.34} &\baseline{30.96} &\baseline{17.32} &\baseline{12.24} &\baseline{9.31}
&\baseline{30.26}&\baseline{30.51}&\baseline{19.95}&\baseline{16.92}&\baseline{9.89}\\
\shline
\end{tabular}}
\label{tab:How2Sign} 
\end{table*}

\begin{table*}[h]
	\centering    
    	\caption{Comparison of SLT performance on  OpenASL.} 
	\resizebox{0.95\textwidth}{!}{
    \begin{tabular}{l|ccccc|ccccc} 
    \shline
\multirow{3}*{Gloss-free SLT}& \multicolumn{10}{c}{OpenASL}   \\
%\cline{2-11}
&\multicolumn{5}{c|}{DEV} & \multicolumn{5}{c}{TEST}\\ 
&ROUGE &BLEU1& BLEU2& BLEU3& BLEU4& ROUGE& BLEU1& BLEU2& BLEU3& BLEU4 \\
\shline  
OpenASL~\cite{shi2022open}&25.31&24.35&14.94&10.72&8.39&24.83& 23.87& 14.08& 9.90& 7.54\\
GloFE-VN~\cite{lin2023gloss} &21.37& 21.06& 12.34 &8.68 &6.68 &21.75& 21.56& 12.74& 9.05& 7.06  \\ 
MixSignGraph~\cite{gan2025mixsigngraph}&25.41&26.82&16.70&11.48&8.36&25.71&26.65 &16.55&11.68&8.69  \\ 
$C^2$RL~\cite{chen2025c} &-& -& - &- &- &31.36&31.46 &21.85 &16.58 &13.21 \\
\baseline{SignLlama 1B} &\baseline{35.28}& \baseline{35.11}& \baseline{26.44}& \baseline{19.02}& \baseline{16.59}& \baseline{37.74} &\baseline{37.37}&\baseline{25.97 }&\baseline{17.70 }&\baseline{15.16}\\  
\shline
	\end{tabular}}  
	\label{tab:OpenASL}  
\end{table*}

\section{Comparisons} 
{For a fair comparison, we only compare with \textbf{image-based GFSLT methods} that do not use \textbf{external SL datasets} for pretraining. }

\paragraph{Evaluation on Phoenix14T Dataset.} As shown in Table~\ref{tab:SLTPhoenix14T}, we compare the GFSLT performance of our model with existing approaches on the Phoenix14T dataset. Our model achieves excellent results, surpassing most of previous methods and improving the BLUE4 score by 1.28 points over SignDINO~\cite{ganlearning}, reaching a 52.81 ROUGE-L score and 26.74 BLEU-4 score on the test set.

\paragraph{Evaluation on CSL-Daily Dataset.}
As shown in Table~\ref{tab:SLTCSL-Daily}, we also evaluate the GFSLT performance of our model on the CSL-Daily dataset. To better support Chinese translation, we replace the original Llama with the Chinese-Llama-2-1.3B model. As shown in the table, our model achieves strong performance with a BLEU-4 score of 24.76, achieving competitive performance.

\paragraph{Evaluation on How2Sign Dataset.} How2Sign is a larger American SL dataset, only provides text labels.  As shown in Table~\ref{tab:How2Sign}, we compare the GFSLT performance of our proposed model with existing models. Our method, incorporating FPG-CTC and VPD, achieves a BLEU4 score of 14.47 on the test set, outperforming the previous SOTA model.

\paragraph{Evaluation on OpenASL Dataset.}Similar to How2Sign, the OpenASL dataset is a large-scale benchmark with a large vocabulary and text-only annotations. Our model achieves excellent performance across all evaluation metrics on both the development and test sets. Compared to the previous best-performing model, C2RL~\cite{chen2025c}, our model achieves substantial improvements, particularly in BLEU scores. On the test set, it improves BLEU-4 from 13.21 to 15.16, and BLEU-1 from 31.46 to 37.37, indicating more accurate and fluent generation.

\section{Conclusion}
\label{sec:conclusion}

In this paper, we focus on fine-tuning large language models (LLMs) for gloss-free sign language translation (GFSLT). We reveal two key challenges in this setting: (1) the inherent distributional gap between SL video features and textual features, which makes alignment difficult; and (2) the common practice of simply concatenating visual and textual inputs in autoregressive training, which causes the model to overemphasize textual cues, deprioritize visual information, and further amplify exposure bias during inference. To address these challenges, we propose two key contributions:  Filtered Pseudo-Gloss CTC Pretraining (FPG-CTC), which leverages automatically generated pseudo-gloss sequences to supervise the training of the visual backbone; and   Visual-Prioritized Distillation (VPD), a training strategy designed to encourage prioritizing visual features. Extensive experiments across multiple benchmark datasets demonstrate the effectiveness of our approach. Our model achieves competitive  results on the gloss-free SLT task.  

\section{Limitations And Discussions} 
Here, we list some potential ideas that can be further explored to improve performance. 
First, our FPG-CTC approach generates pseudo-gloss sequences using simple rule-based NLP techniques. More discussion of effective pseudo-gloss generation strategies  can be further investigated in future work.
Second,  VPD requires dual forward passes for visual-text prediction and visual-only prediction, increasing training cost. More efficient approaches to enhance visual feature utilization are worth investigating. In addition, we highlight the potential negative social impacts.
Our model, like other deep learning approaches, requires significant computational resources, which could raise issues related to energy use and environmental impact. Moreover, since our method is data-driven, it may inherit potential biases from the training data. Careful dataset selection and balanced data are important to reduce such risks.

\paragraph{Scaling to Larger Models.} Due to hardware limitations, we scale the LLM from 1B up to 13B in our experiments. In future work, we plan to explore larger pretrained LLMs, as emergent capabilities observed at greater scales may lead to further gains in GFSLT.

\begin{acks}
This work is supported in part by National Natural Science Foundation of China under Grant Nos. 62172208, 92467202, 62272216; Key Projects of Jiangsu Provincial Basic Research Program under Grant No. BK20243040; JiangSu Natural Science Foundation under Grant No. BK20251989. This work is partially supported by Fundamental and Interdisciplinary Disciplines Breakthrough Plan of the Ministry of Education of China (No. JYB2025XDXM118); the “111 Center” (No. B26023); Collaborative Innovation Center of Novel Software Technology and Industrialization.
\end{acks}

\bibliographystyle{ACM-Reference-Format}
\balance
\bibliography{bib/slt, bib/cslr, bib/islr, bib/other}

\newpage
\appendix
\clearpage

\setcounter{table}{10}
\setcounter{figure}{4}

\section{Datasets}
\paragraph{Details of the Datasets.}
We provide detailed information on the five datasets used in our paper in Table~\ref{tab:dataset}. The datasets we downloaded may slightly differ from the official versions described in their respective papers. We also provide the token vocabulary obtained by the Llama tokenizer and the vocabulary in FPG-CTC of processed text for reference.

\iffalse
\begin{figure}
    \centering
\includegraphics[width=0.9\linewidth]{Section/figure/fpgctc.pdf}
\caption{The Processing of FPG-CTC.}
\vspace{-4mm}
\label{fig:FPG-CTC}  
\end{figure}
\fi

\section{Model Details}
Our FPG-CTC module generates a pseudo-gloss sequence from the target text by applying lemmatization and removing function words such as prepositions and conjunctions. Since certain words in the text may not be explicitly expressed in the corresponding SL video, we further introduce random dropping to remove a fixed proportion of tokens, producing the final pseudo-gloss labels. The NLTK toolkit is used for pseudo-gloss preprocessing steps.

\section{Ablation Study on Hyperparameters}

\paragraph{Effect of Different $\lambda$.} In the main paper, we illustrated the impact of different $\lambda$ values on our model’s performance using a line chart (Figure~\ref{fig:droprate}). Here, we provide the detailed results in Table~\ref{tab:vpd_weight}. As shown, when $\lambda=0$ (i.e., without the VPD loss), the model achieves the lowest performance. Increasing $\lambda$ leads to notable improvements across ROUGE and BLEU metrics, peaking around $\lambda=5$ to $\lambda=10$. Therefore, we set $\lambda$ to 5 in our final model to balance performance and stability.

\paragraph{Effect of Different Drop Rates.} In the main paper, we illustrated the impact of different drop rates in FPG-CTC using a line chart (Figure~\ref{fig:droprate}). Here, we provide detailed quantitative results in Table~\ref{tab:droprate}. The model performance improves as the drop rate increases from 0 to 0.2, reaching the best results at a drop rate of 0.2 across all ROUGE and BLEU metrics. However, further increasing the drop rate beyond 0.2 leads to significant performance degradation, as excessive word dropping hinders effective alignment learning.

\begin{table*}[t]
	\centering 
    \caption{Details of datasets used in our paper. \textbf{Voc}: vocabulary size. \textbf{PG}: pseudo gloss.} 
	\resizebox{0.7\textwidth}{!}{
\begin{tabular}{l|c|c|rrr} 
\shline
\multirow{2}*{Dataset}&\multirow{2}*{Token Voc in Text} &\multirow{2}*{PG Voc in FPG-CTC}&\multicolumn{3}{|c}{Video Samples} \\ 
\cline{4-6}
&& &  {Train}&{Test}&{Validation}  \\ 
\toprule 
Phoenix14T        &2121 &30,01 &7,096&642&519 \\
CSL-Daily &7277&  2377& 18,401&1,176&1,077\\ 
How2Sign  &9534& 15347&30,904&2328&1713\\ 
OpenASL&19,250&25,402&96,476&975&966\\
\shline
	\end{tabular}}   
	\label{tab:dataset}
\end{table*}

\begin{table*}[t] 
	\centering 
	\caption{Effect of different $\lambda$ in Equation 7.}
	\resizebox{0.87\linewidth}{!}{ 
		\begin{tabular}{c|ccccc|ccccc}
		\shline 
\multirow{2}*{$\lambda$}& \multicolumn{5}{c|}{DEV}& \multicolumn{5}{c}{TEST} \\
&ROUGE &BLEU1& BLEU2& BLEU3& BLEU4& ROUGE& BLEU1& BLEU2& BLEU3& BLEU4 \\
			\shline 
0 &50.96 &50.68 & 35.98 & 28.75&23.83 &51.08& 50.16&38.08&29.51&23.52	 \\
1&{51.03}  &{51.73}&{38.07} &{29.26} &{25.15} &{51.41} &{50.50} &{39.16} &{31.27} &{25.04}  \\
5 &\better{52.53}  &\better{52.13}&\better{38.88} &\better{30.61} &\better{25.73} &\better{52.81} &\better{52.54} &\better{40.96} &\better{32.87} &\better{26.74}\\
10 &{52.10} &{51.21} &{38.79} &{29.97} &{25.70} &{52.61} &{51.91} &{39.72} &{31.61} &{26.20}	 \\
15 &51.13 &{51.16} &{38.65} &{29.77} &{25.54} &51.31 &49.87 &37.31 &29.32 &26.07	 \\
20 &{51.71} &51.66 &38.36 &29.52 &25.27 &51.48 &49.75 &37.29 &29.33 &26.05 \\
25 &51.57 &51.87 &38.51 &29.62 &25.35 &51.20 &50.11 &37.58 &29.58 &{26.28}\\
	\shline 
	\end{tabular} }
	\label{tab:vpd_weight} 
\end{table*}

\begin{table*}[t] 
	\centering
	\caption{Effect of different drop rate in FPG-CTC.}
	\resizebox{0.87\linewidth}{!}{ 
		\begin{tabular}{l|ccccc|ccccc|cc|cc}
		\shline 
\multirow{2}*{DropRate}& \multicolumn{5}{c|}{DEV}& \multicolumn{5}{c|}{TEST} &\multicolumn{2}{c|}{Dev}& \multicolumn{2}{c}{Test} \\
&ROUGE &BLEU1& BLEU2& BLEU3& BLEU4& ROUGE& BLEU1& BLEU2& BLEU3& BLEU4  &WER&Del/Ins&WER&Del/Ins\\
			\shline
0 & 48.77&47.53&34.57 &26.56&23.40&49.05&48.28&35.59&27.68&24.56 &69.55& 53.67/ 5.23& 69.35& 53.76/ 4.65\\
0.1 & \better{50.35} &\better{48.63}&
\better{36.03}&\better{28.06} &\better{24.72} &\better{50.31}  &\better{49.39}&\better{36.88} &\better{28.94} &\better{25.74} &72.95& 63.67/ 0.44& 72.77& 63.03/ 0.45\\
0.2 &  \better{52.53}  &\better{52.13}&\better{38.88} &\better{30.61} &\better{25.73} &\better{52.81} &\better{52.54} &\better{40.96} &\better{32.87} &\better{26.74}& {78.77}&74.77/ 0.15& 78.95&73.68/0.18	 \\
0.3& 48.65 &47.62&36.14 &28.02 &22.76 &48.86 &48.33&35.39&27.12 &23.68&88.28& 86.95/ 0.03& 88.36& 86.99/ 0.01\\
0.4&45.80 &44.97&33.65 &25.70 &20.64 & 45.87  &46.13&33.14 &25.25 &22.16 &94.60&94.25/ 0.00&94.12& 93.73/0.01\\
0.5&42.24 &41.76&29.99&22.23&17.53&41.64 &42.75&29.51 &21.73 &18.92 &94.77& 94.43/ 0.00 &94.63&94.17/0.01\\
	\shline 
	\end{tabular} }
	\label{tab:droprate}
\end{table*}

\begin{table*}[t]
\tablestyle{2pt}{1.2} 
	\centering 
    	\caption{Effect of proposed FPG-CTC for SLT. Besides, we also show the CSLR performance based on pseudo gloss sequence obtained by FPG-CTC in the right part.}
	\resizebox{0.97\textwidth}{!}{ 
		\begin{tabular}{c|c|ccccc|ccccc|cc|cc}
\hline
\multirow{2}*{Dataset}&\multirow{2}*{Model} & \multicolumn{5}{c|}{Dev}& \multicolumn{5}{c|}{Test} &\multicolumn{2}{c|}{Dev}& \multicolumn{2}{c}{Test}  \\
&&ROUGE&BLEU1&BLEU2&BLEU3&BLEU4&ROUGE&BLEU1&BLEU2&BLEU3&BLEU4 &WER&Del/Ins&WER&Del/Ins\\
			\shline 
\multirow{2}*{Phoenix14T}
&w/o FPG-CTC &20.11&17.22 &8.21 &6.18&5.23& 20.09 &18.12&58.56&6.46&5.46 &-&-/-&-&-/- \\ 
&w/ FPG-CTC& \better{52.53}  &\better{52.13}&\better{38.88} &\better{30.61} &\better{25.73} &\better{52.81} &\better{52.54} &\better{40.96} &\better{32.87} &\better{26.74}&78.62& 72.57/00.22&78.32&72.23/00.24\\  
 
\hline
\multirow{2}*{CSL-Daily}
&w/o FPG-CTC &36.51& 34.54  &23.72&14.655 &9.11&35.31&33.65&19.34&12.67&8.27 &-&-/-&-&-/- \\ 
&w/ FPG-CTC &\better{51.62}&\better{52.12}&\better{41.23}&\better{33.23}&\better{24.83}&\better{51.43}&\better{52.38}&\better{39.78}&\better{33.91}&\better{24.76}&69.75 &  56.79/1.60&69.77& 55.89/1.79\\  
 
\hline
\multirow{2}*{How2Sign}
&w/o FPG-CTC &19.45 &24.15   &13.56 &7.90 & 5.15&19.15&24.47&12.98&7.81&5.14&-&-/-&-&-/- \\ 
&w/ FPG-CTC&  \better{37.34} &\better{37.96} &\better{22.32} &\better{17.24} &\better{14.31}&\better{39.26}&\better{40.51}&\better{26.95}&\better{18.92}&\better{14.47}&70.62& 69.57/00.22&70.32&68.23/00.24\\  
 
\hline
\multirow{2}*{OpenASL}
&w/o FPG-CTC &13.13 &12.69 &4.57 &2.39 &1.96&12.68 &11.27&4.87&2.45&1.98&-&-/-&-&-/- \\ 
&w/ FPG-CTC&\better{35.28}& \better{35.11}& \better{26.44}& \better{19.02}& \better{16.59}& \better{37.74} &\better{37.37}&\better{25.97 }&\better{17.70 }&\better{15.16}&82.57&69.21/0.65&82.85&69.19/0.69\\  
\hline
	\end{tabular}} 
	\label{tab:TCP_other}
\end{table*}

\begin{figure*}[t]
\centering            
\includegraphics[width=0.9\textwidth]{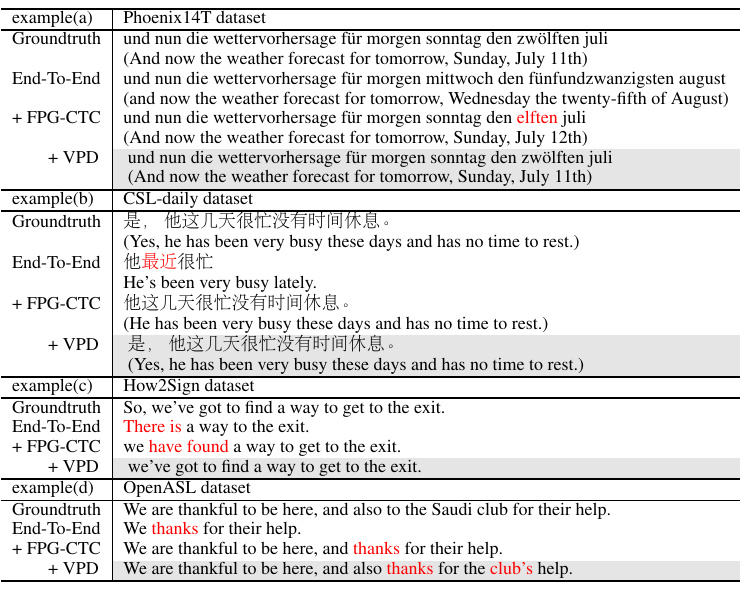} 
\caption{SLT  Qualitative results on Phoenix-2014T, CSL-Daily, How2Sign and OpenASL.}
\label{fig:SLTQualitative} 
\end{figure*}

\section{Ablation Study on Proposed Module}
\paragraph{Effect of  FPG-CTC.} 
We evaluate the effectiveness of our proposed FPG-CTC module by comparing the SLT performance on four benchmark datasets with and without FPG-CTC pretraining. As shown in Table~\ref{tab:TCP_other}, incorporating FPG-CTC consistently yields significant improvements across all datasets and evaluation metrics. For instance, on the Phoenix14T dataset, ROUGE and BLEU-4 scores improve from 20.09/5.46 to 52.81/26.74 on the test set, indicating that the pseudo-gloss supervision helps align sign video features with text features more effectively.

Additionally, we report the CSLR performance based on the pseudo-gloss sequences generated by FPG-CTC (right side of Table~\ref{tab:TCP_other}). Although the pseudo-gloss labels are not ground truth and the resulting WERs are suboptimal, they still demonstrate a reasonable level of recognition performance, suggesting that the pseudo sequences preserve coherent structure and temporal alignment. Overall, these results demonstrate that FPG-CTC offers strong supervisory signals for improving gloss-free SLT, and lays a solid foundation for high-quality translation without relying on ground-truth gloss annotations.

\section{Visualization}

\paragraph{SLT Qualitative Results.}  
Figure~\ref{fig:SLTQualitative}  presents a qualitative analysis of our SignLlama model on the gloss-free SLT task, with examples selected from the test sets of Phoenix14T, CSL-Daily, How2Sign, and OpenASL.
The results show that SignLlama, when equipped with both FPG-CTC and VPD, produces the most accurate translations. In contrast, the end-to-end fine-tuning model performs worse, highlighting the effectiveness of our proposed FPG-CTC and VPD.

\begin{figure*}[t]
    \centering
    \begin{subfigure}[b]{0.24\textwidth}
        \includegraphics[width=\linewidth]{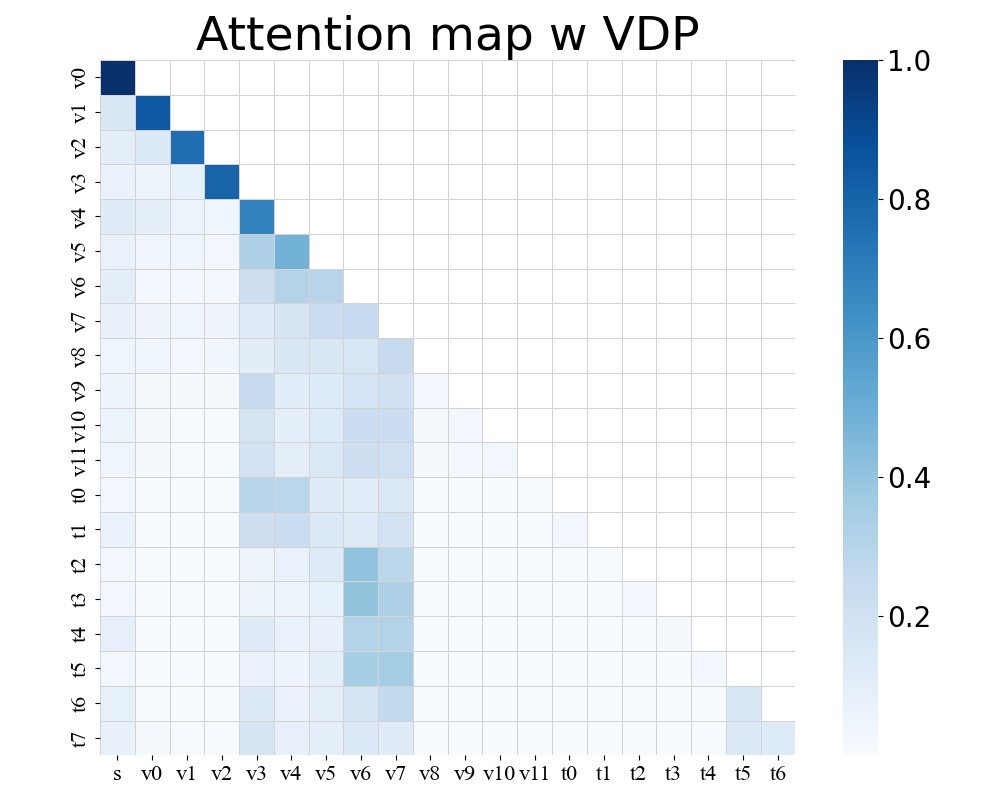}
        \caption{head=0} 
    \end{subfigure}
       \begin{subfigure}[b]{0.24\textwidth}
        \includegraphics[width=\linewidth]{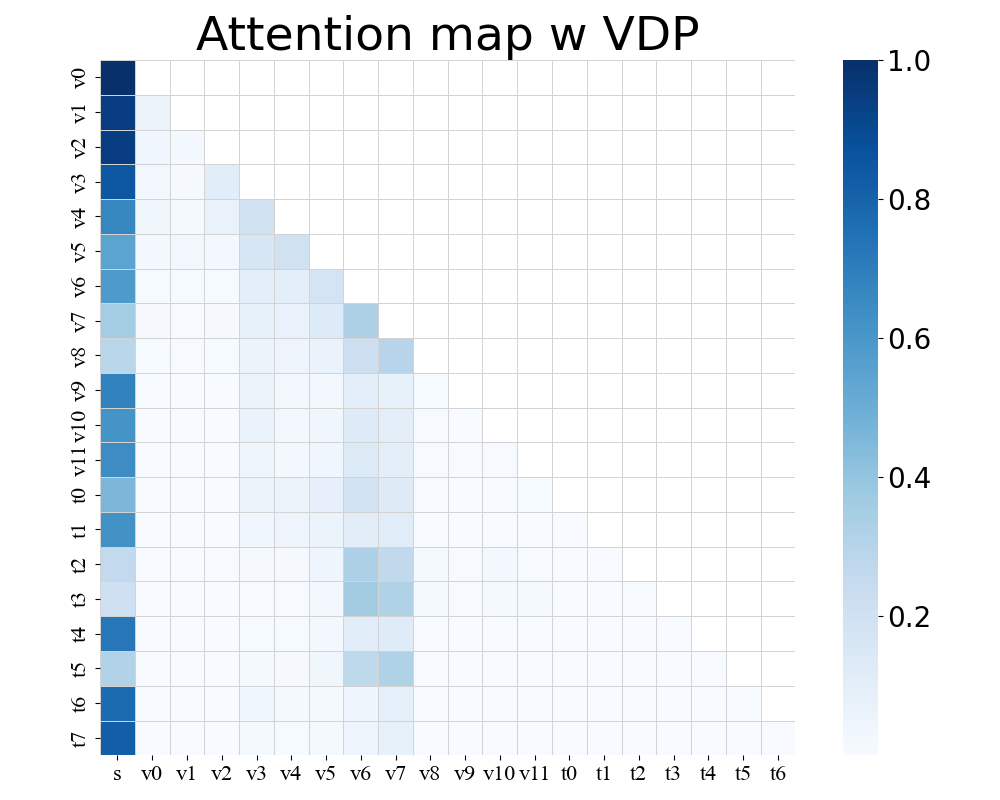}
        \caption{head=2} 
    \end{subfigure}
       \begin{subfigure}[b]{0.24\textwidth}
        \includegraphics[width=\linewidth]{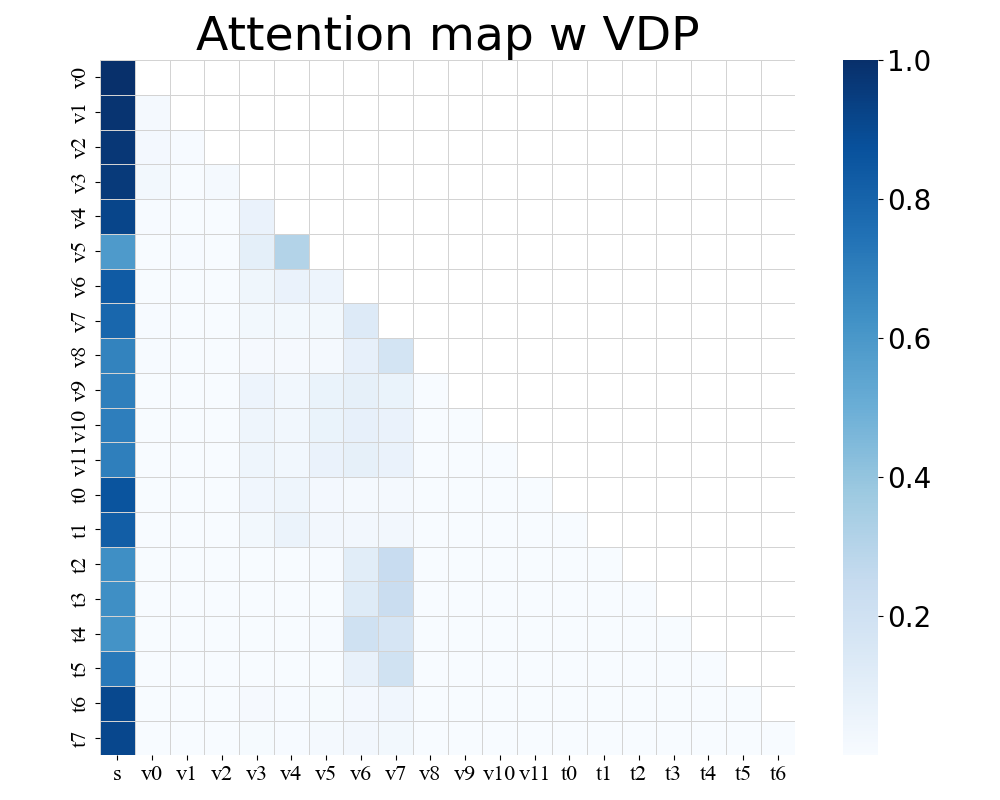}
        \caption{head=4} 
    \end{subfigure}
       \begin{subfigure}[b]{0.24\textwidth}
        \includegraphics[width=\linewidth]{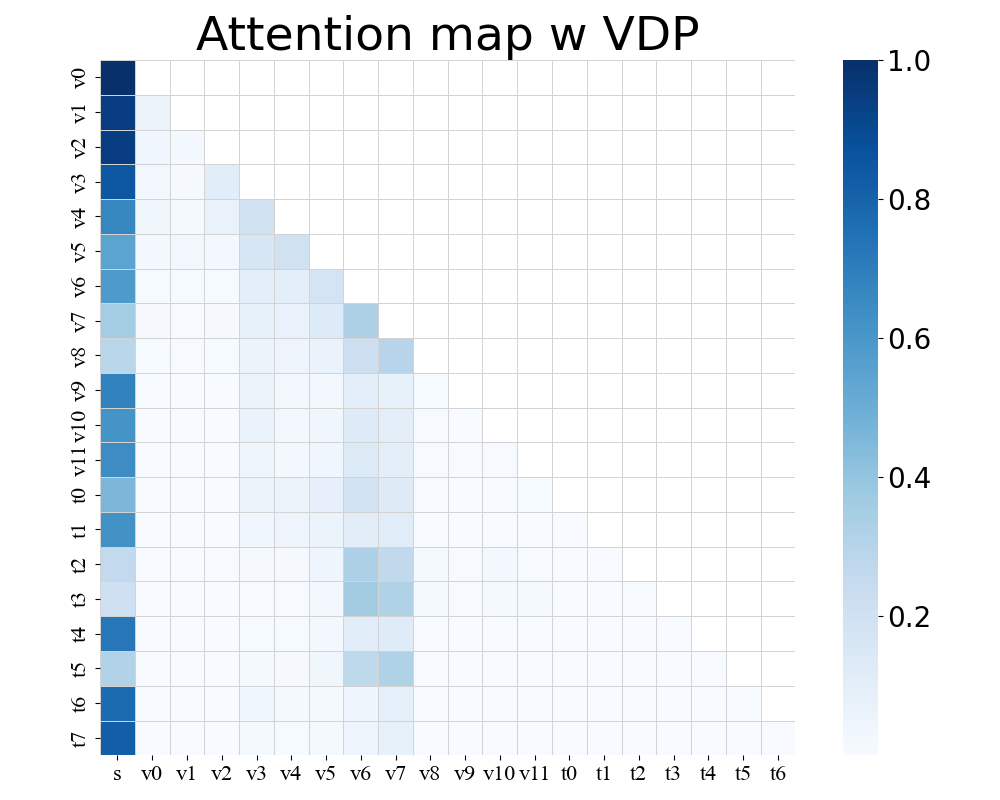}
        \caption{head=6} 
    \end{subfigure}
       \begin{subfigure}[b]{0.24\textwidth}
        \includegraphics[width=\linewidth]{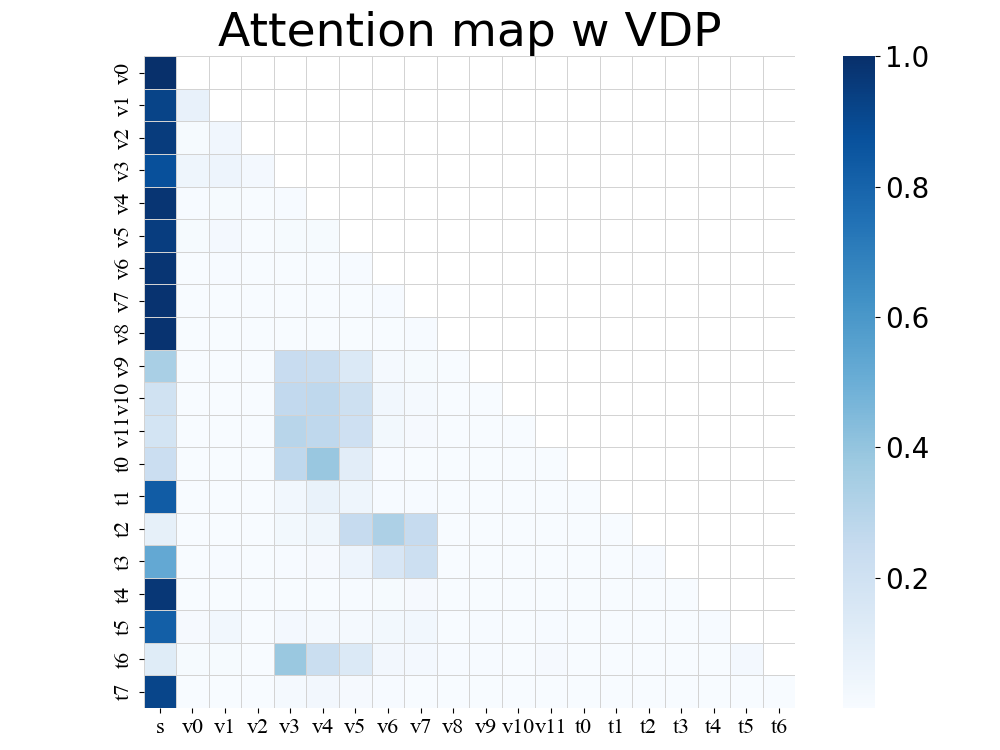}
        \caption{head=7} 
    \end{subfigure}
       \begin{subfigure}[b]{0.24\textwidth}
        \includegraphics[width=\linewidth]{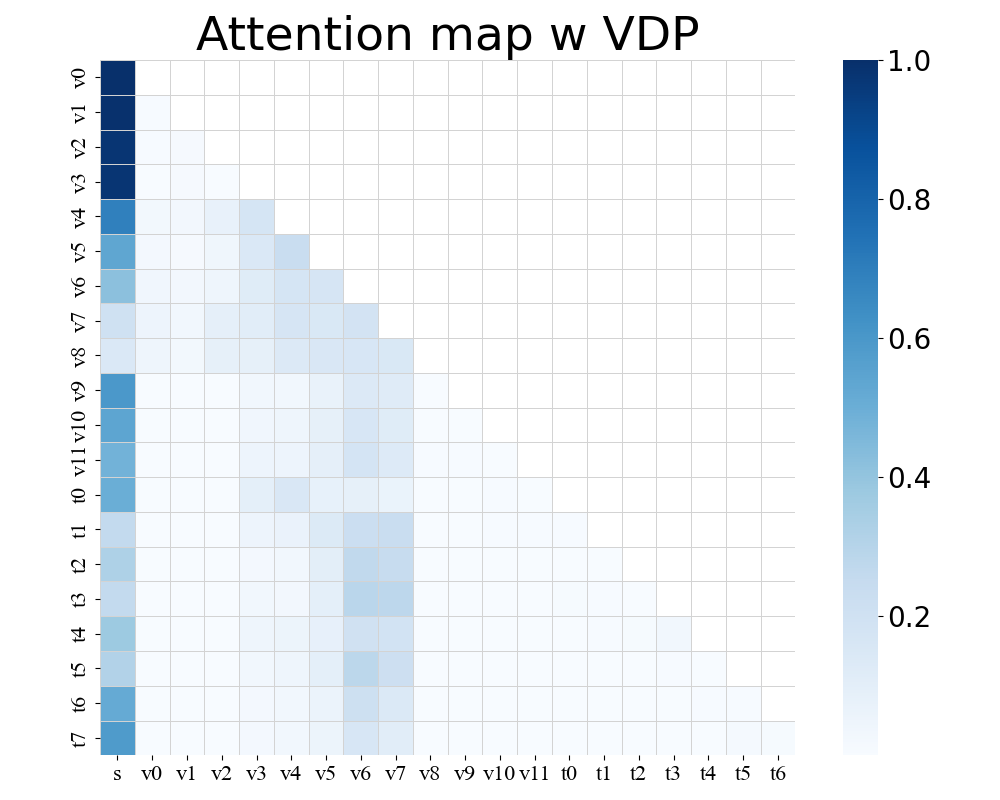}
        \caption{head=8}
    \end{subfigure}
     \begin{subfigure}[b]{0.24\textwidth}
        \includegraphics[width=\linewidth]{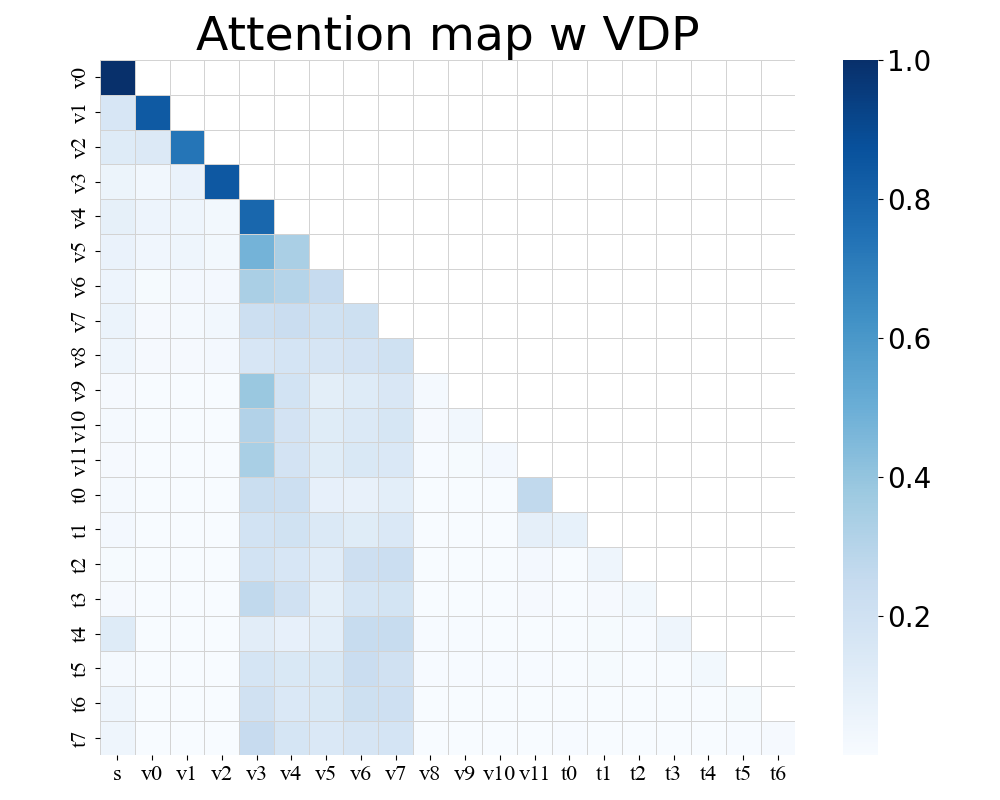}
        \caption{head=15}
    \end{subfigure}
           \begin{subfigure}[b]{0.24\textwidth}
        \includegraphics[width=\linewidth]{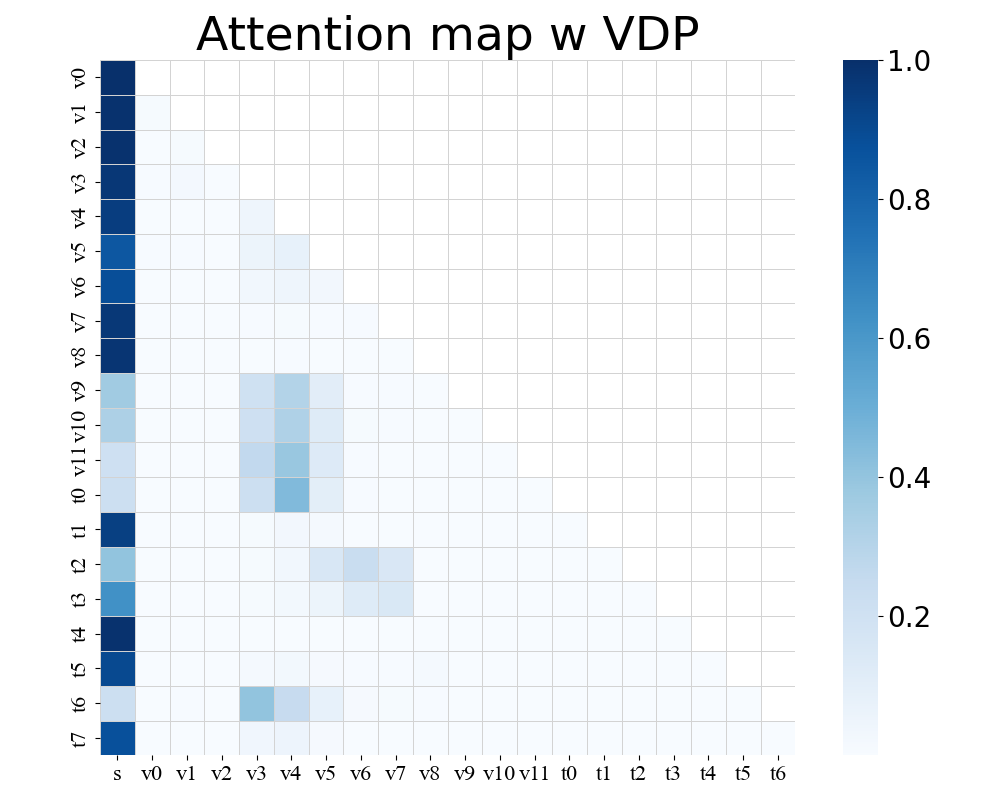}
        \caption{head=18}
    \end{subfigure}
           \begin{subfigure}[b]{0.24\textwidth}
        \includegraphics[width=\linewidth]{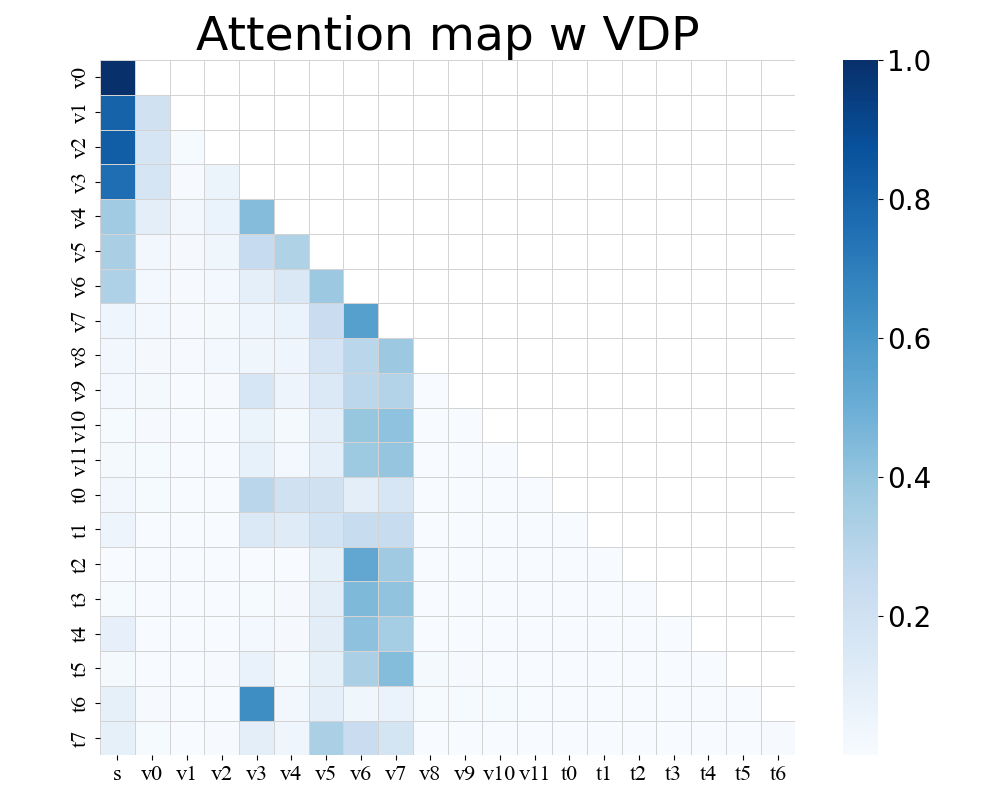}
        \caption{head=20}
    \end{subfigure}
    \begin{subfigure}[b]{0.24\textwidth}
        \includegraphics[width=\linewidth]{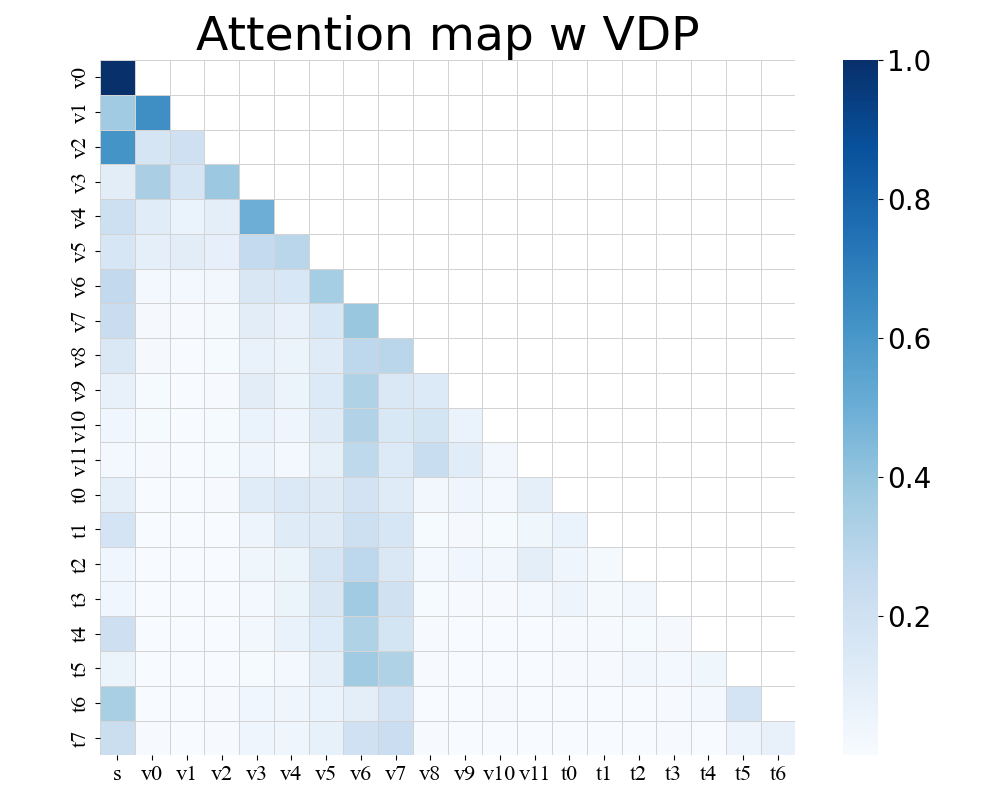}
        \caption{head=24}
    \end{subfigure}
               \begin{subfigure}[b]{0.24\textwidth}
        \includegraphics[width=\linewidth]{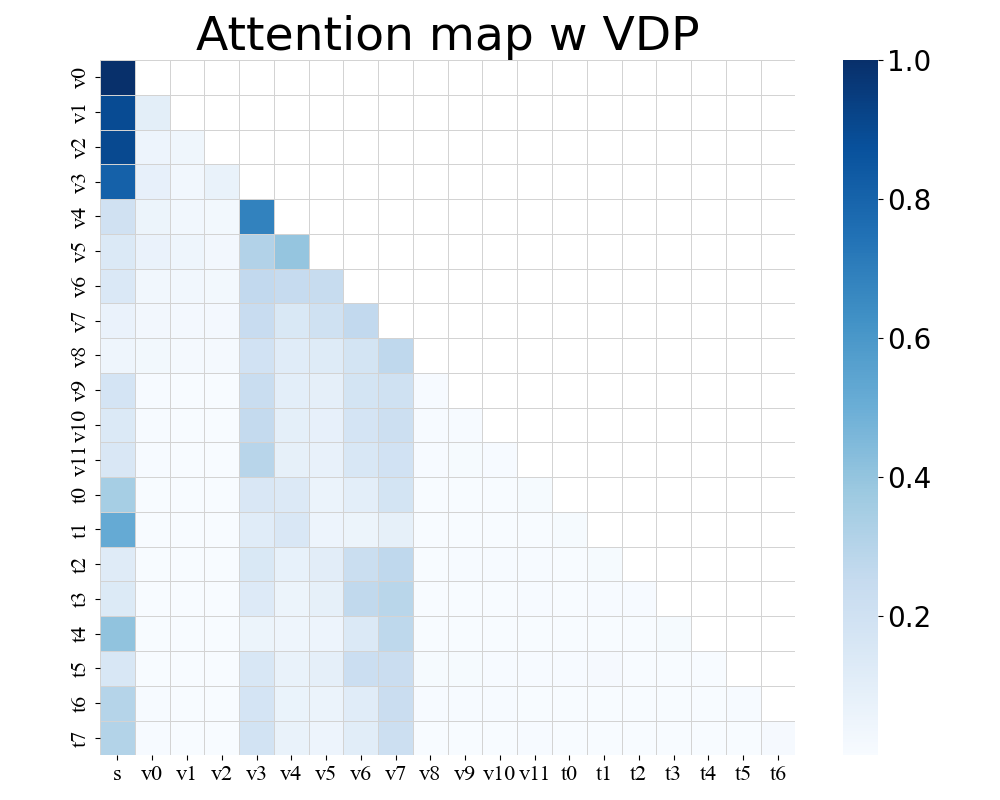}
        \caption{head=28}
    \end{subfigure}
               \begin{subfigure}[b]{0.24\textwidth}
        \includegraphics[width=\linewidth]{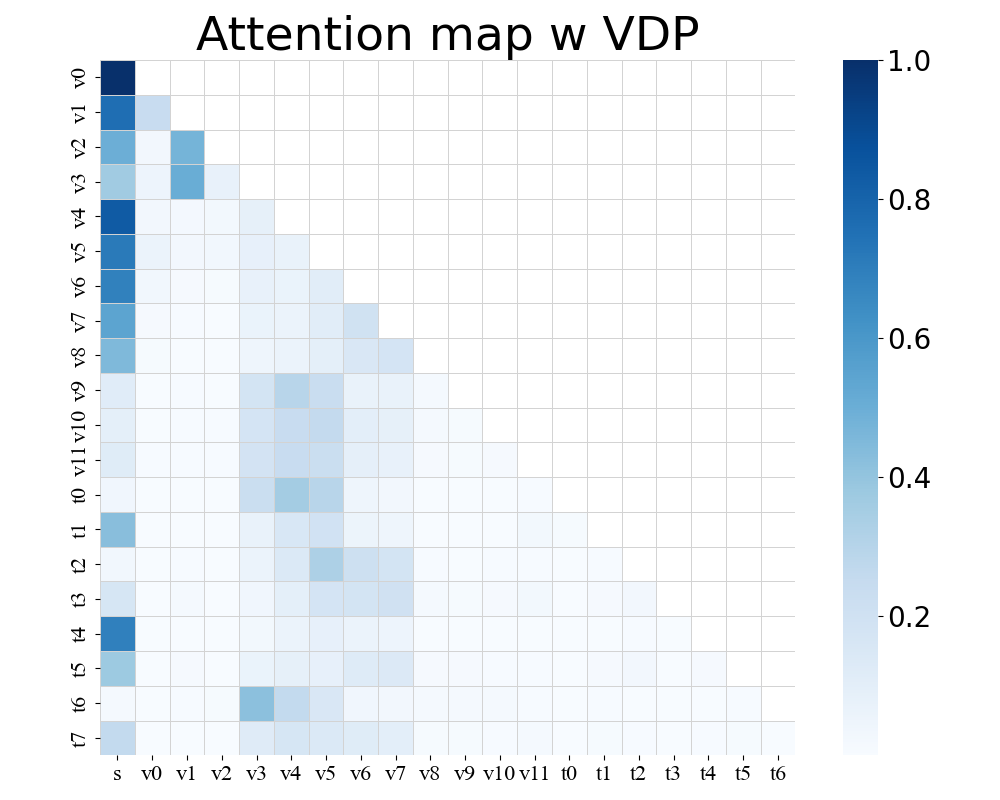}
        \caption{head=30}
    \end{subfigure}
    \caption{Visualization of attention maps in the last layer of Llama \textbf{with}  VPD, showing that the attention is mainly focused on visual tokens. }
    \label{fig:attn_maps}
\end{figure*}

\begin{figure*}[t]
    \centering
    \begin{subfigure}[b]{0.24\textwidth}
        \includegraphics[width=\linewidth]{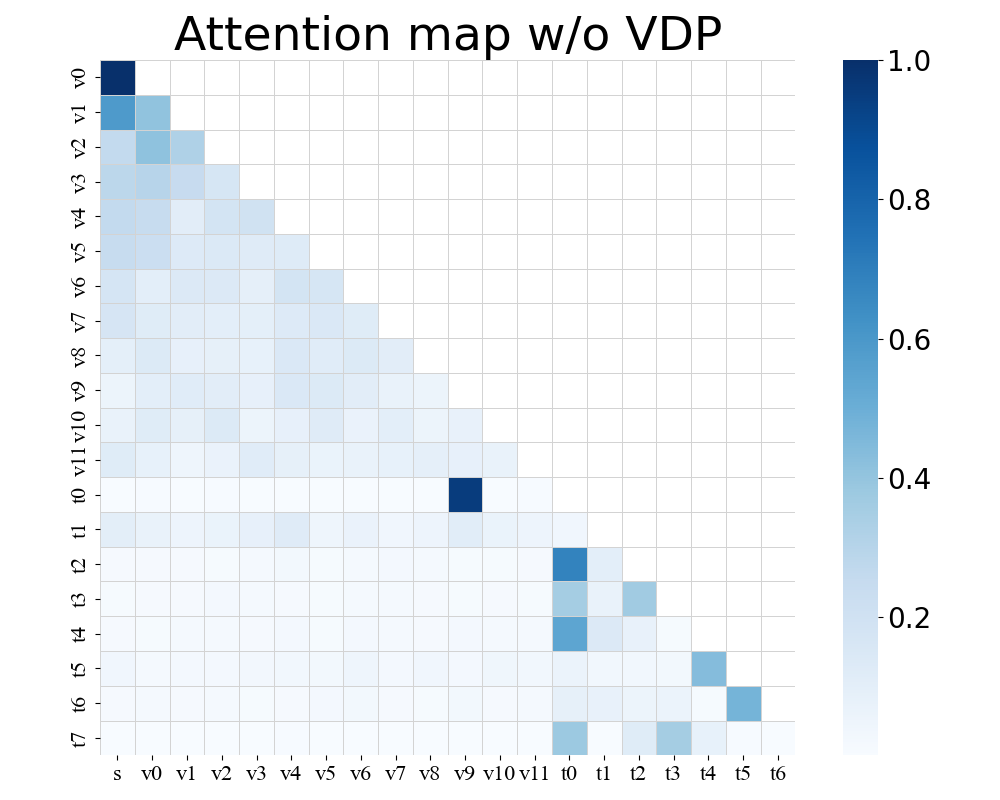}
        \caption{head=0} 
    \end{subfigure}
       \begin{subfigure}[b]{0.24\textwidth}
        \includegraphics[width=\linewidth]{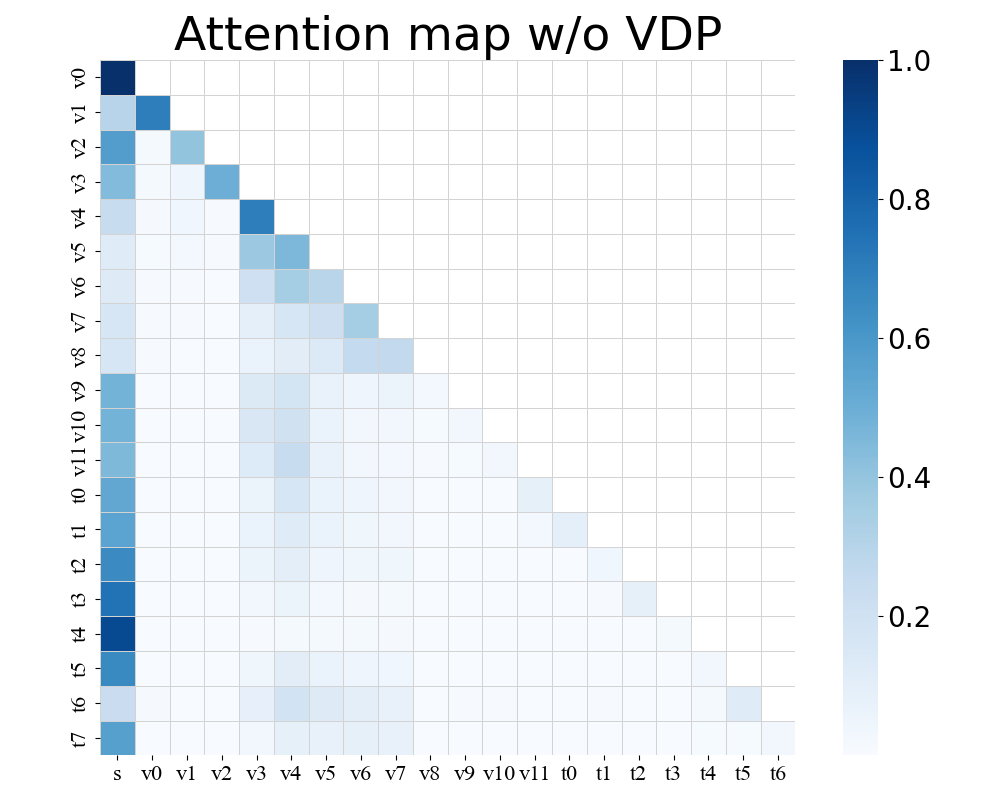}
        \caption{head=2} 
    \end{subfigure}
       \begin{subfigure}[b]{0.24\textwidth}
        \includegraphics[width=\linewidth]{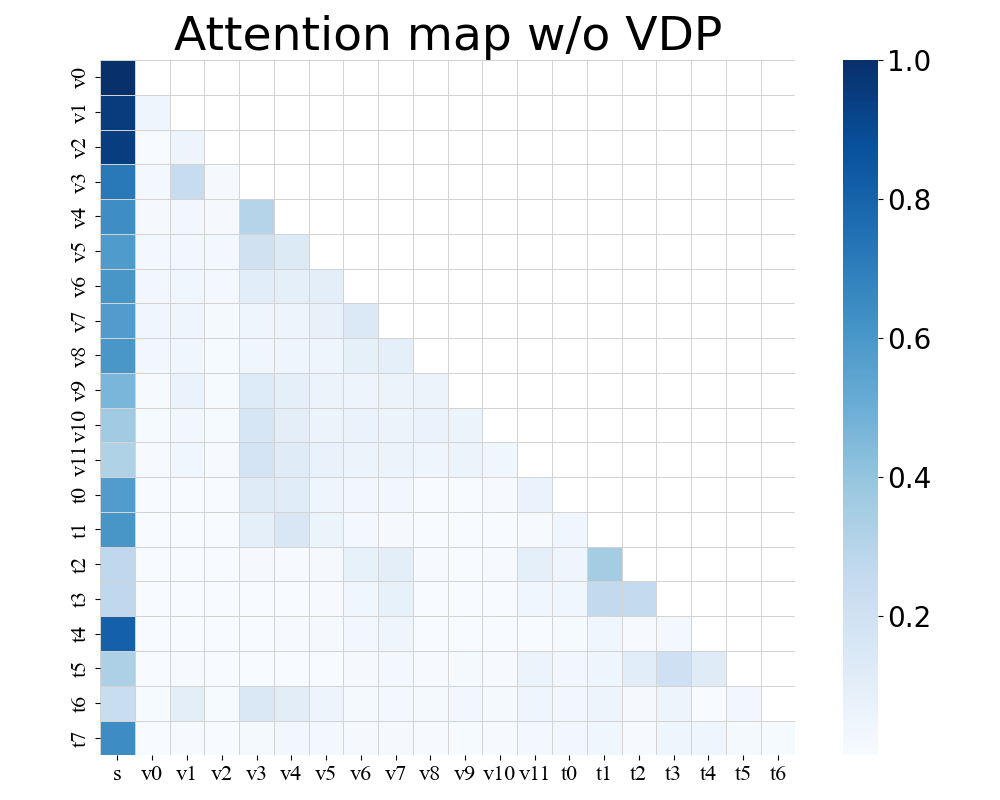}
        \caption{head=4} 
    \end{subfigure}
       \begin{subfigure}[b]{0.24\textwidth}
        \includegraphics[width=\linewidth]{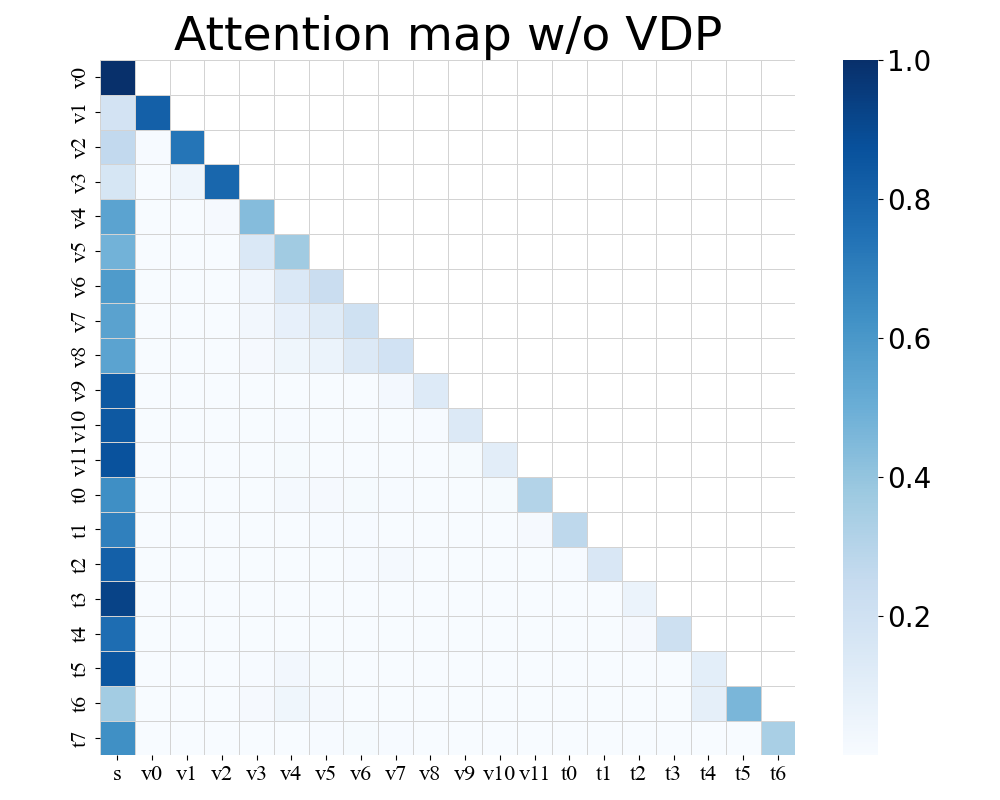}
        \caption{head=6} 
    \end{subfigure}
       \begin{subfigure}[b]{0.24\textwidth}
        \includegraphics[width=\linewidth]{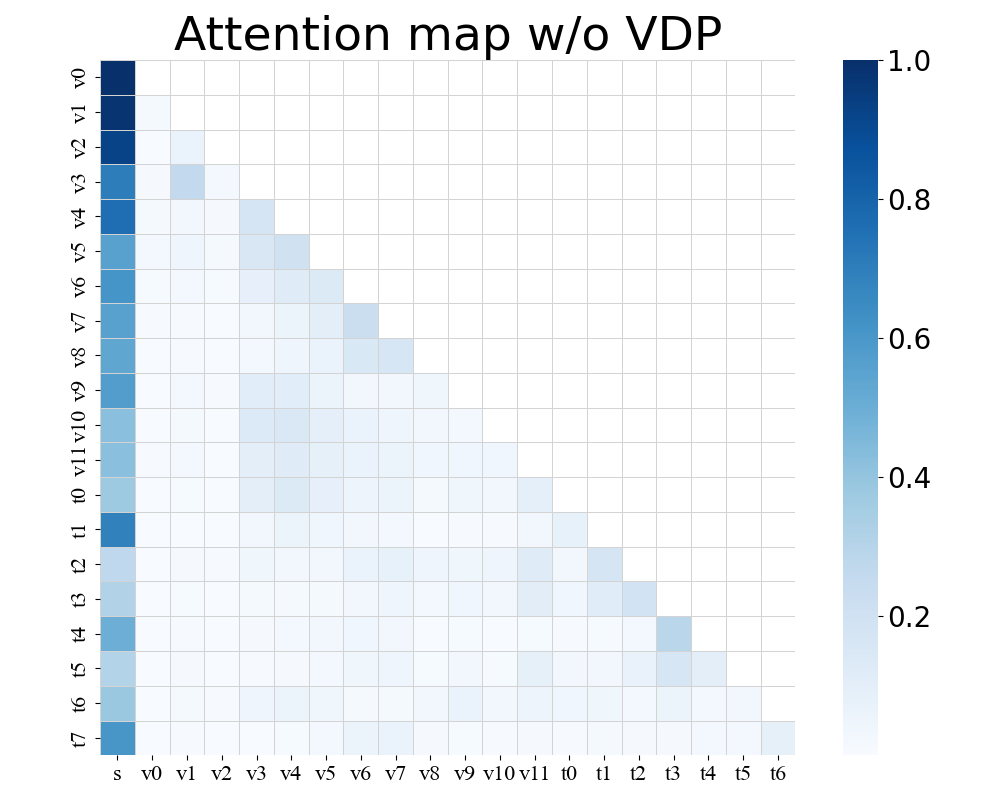}
        \caption{head=7} 
    \end{subfigure}
       \begin{subfigure}[b]{0.24\textwidth}
        \includegraphics[width=\linewidth]{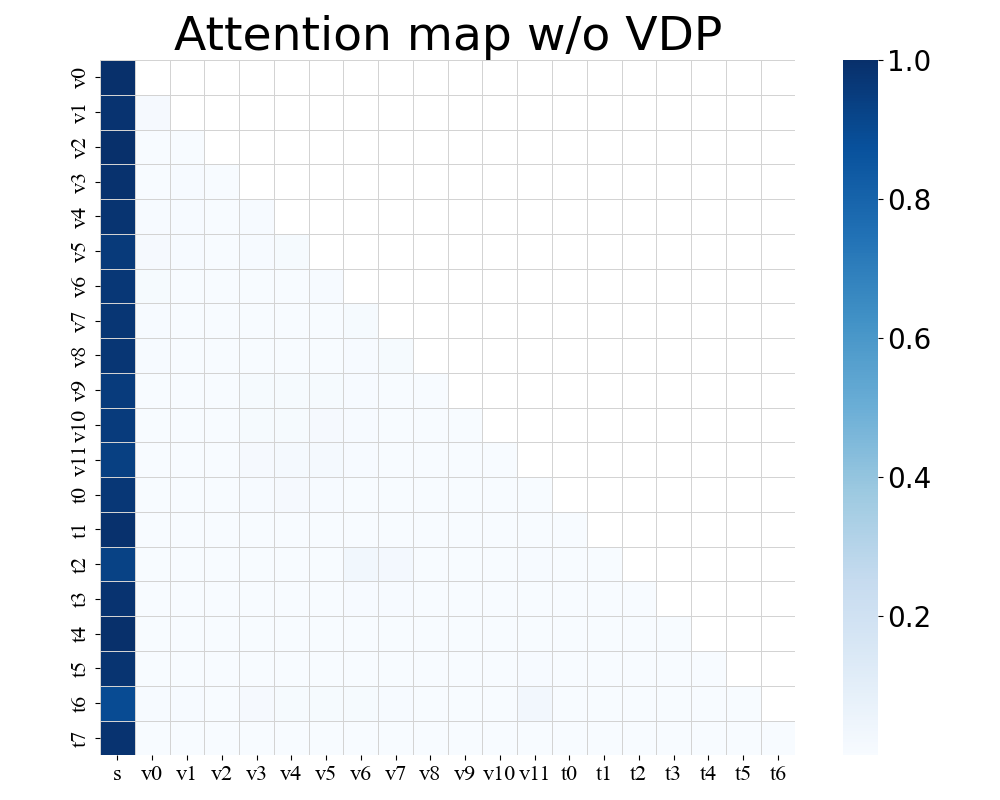}
        \caption{head=8}
    \end{subfigure}
     \begin{subfigure}[b]{0.24\textwidth}
        \includegraphics[width=\linewidth]{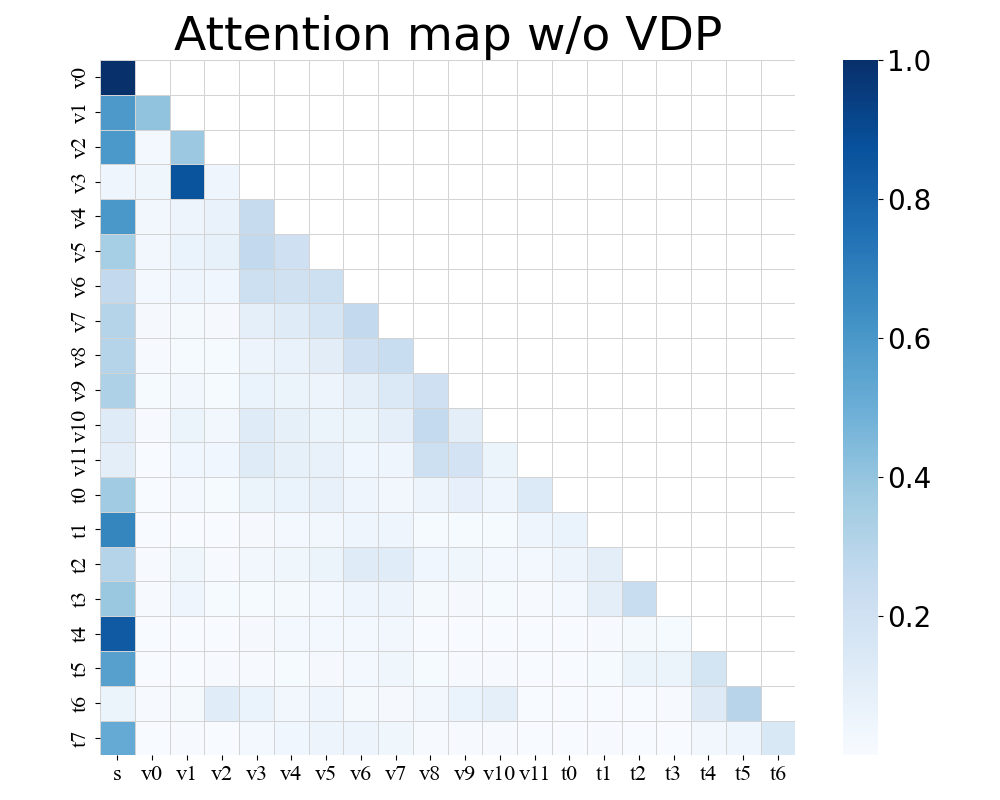}
        \caption{head=15}
    \end{subfigure}
           \begin{subfigure}[b]{0.24\textwidth}
        \includegraphics[width=\linewidth]{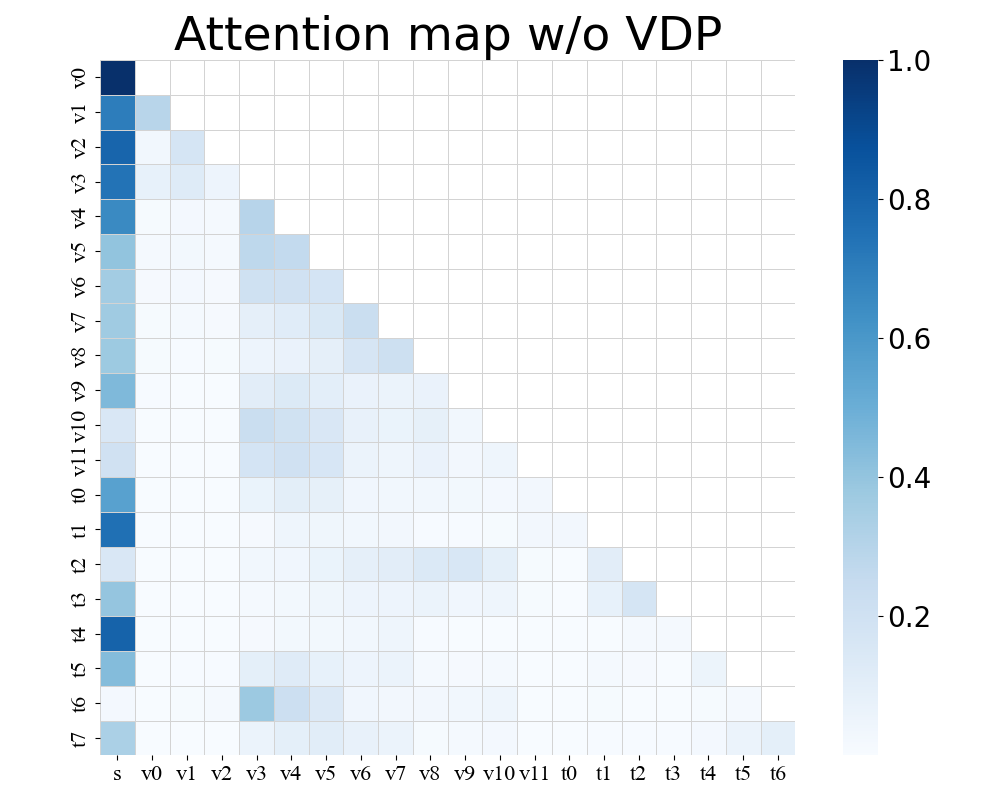}
        \caption{head=18}
    \end{subfigure}
           \begin{subfigure}[b]{0.24\textwidth}
        \includegraphics[width=\linewidth]{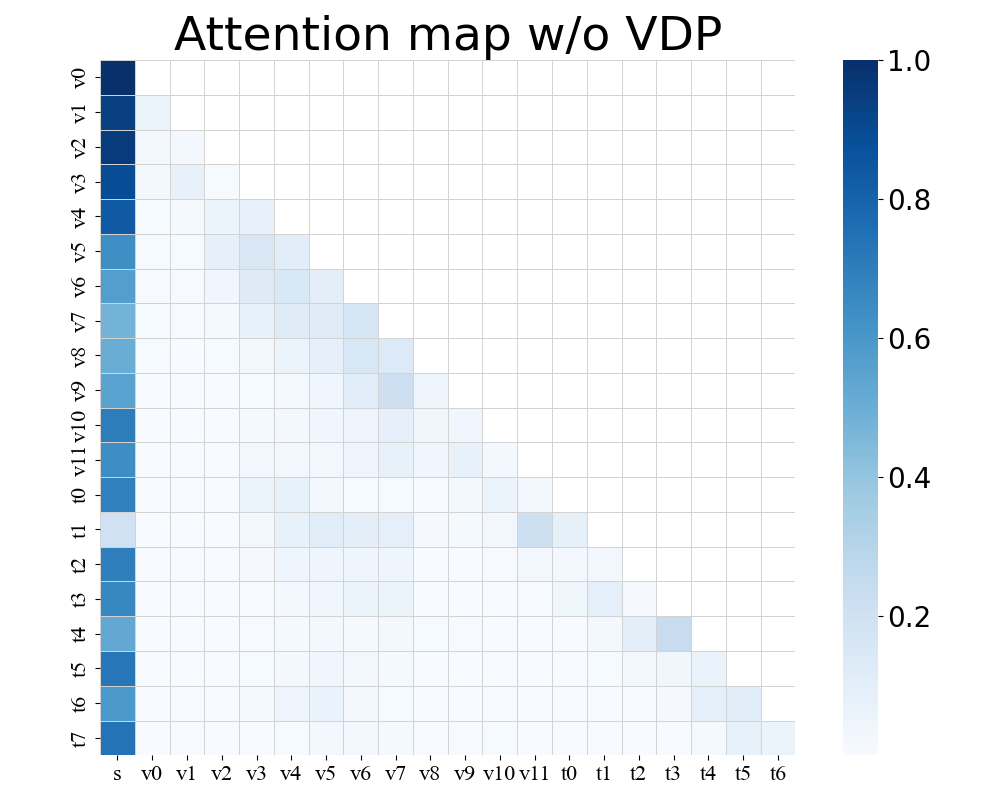}
        \caption{head=20}
    \end{subfigure}
    \begin{subfigure}[b]{0.24\textwidth}
        \includegraphics[width=\linewidth]{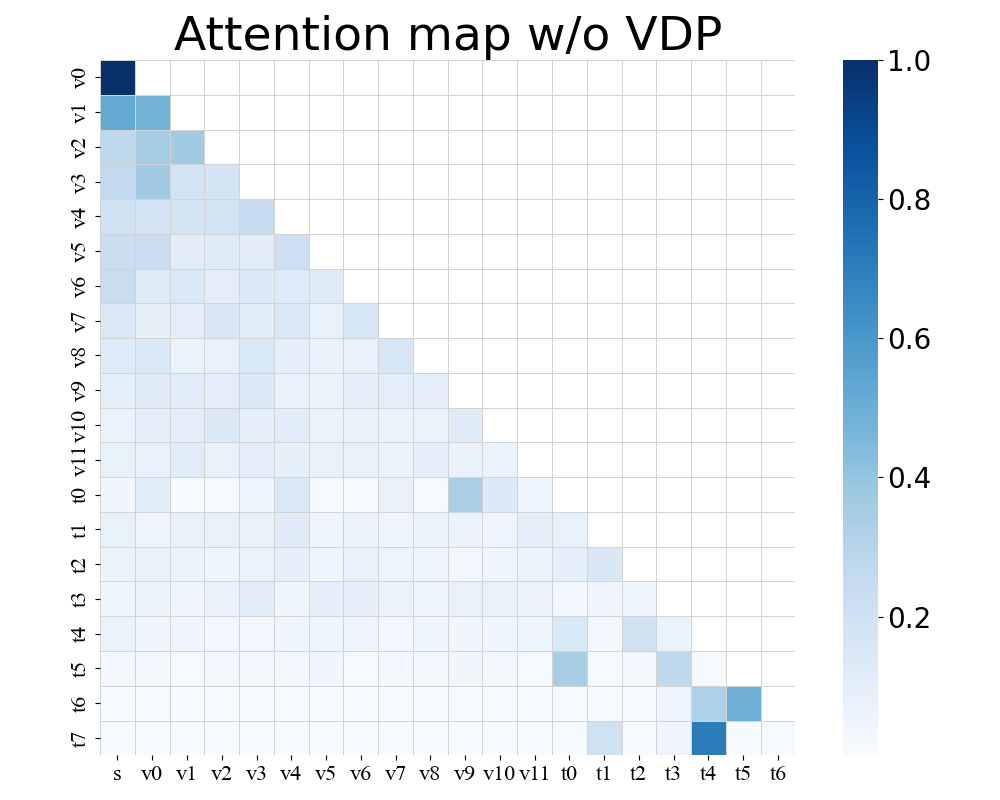}
        \caption{head=24}
    \end{subfigure}
               \begin{subfigure}[b]{0.24\textwidth}
        \includegraphics[width=\linewidth]{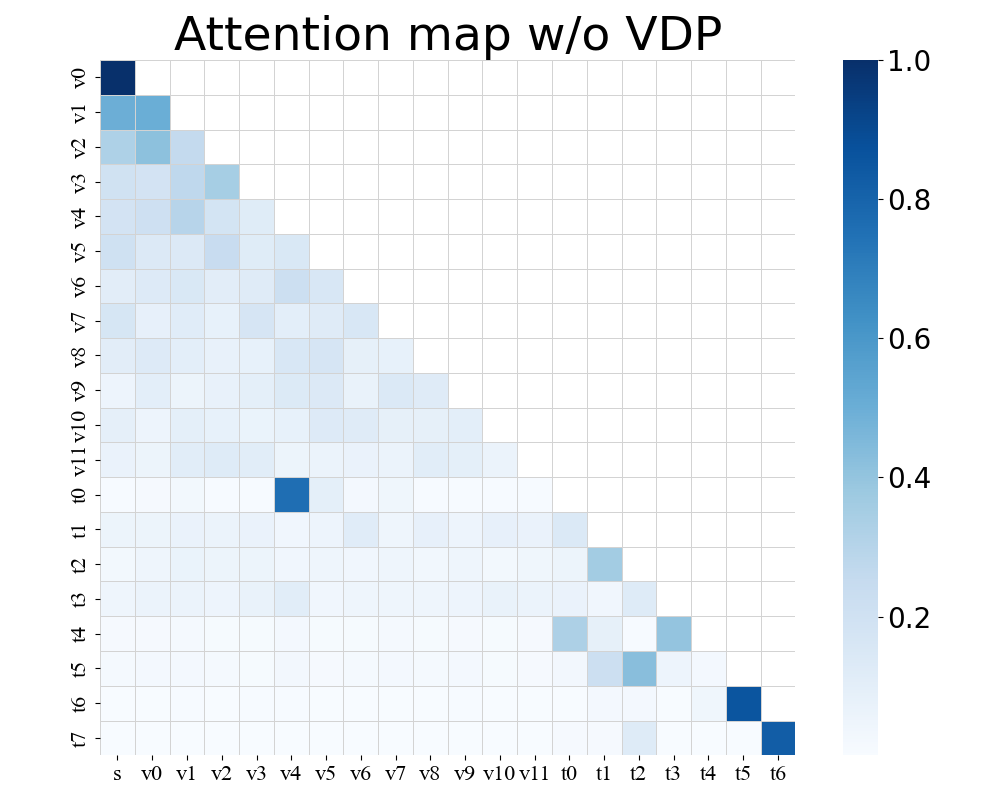}
        \caption{head=28}
    \end{subfigure} 
    \begin{subfigure}[b]{0.24\textwidth}
        \includegraphics[width=\linewidth]{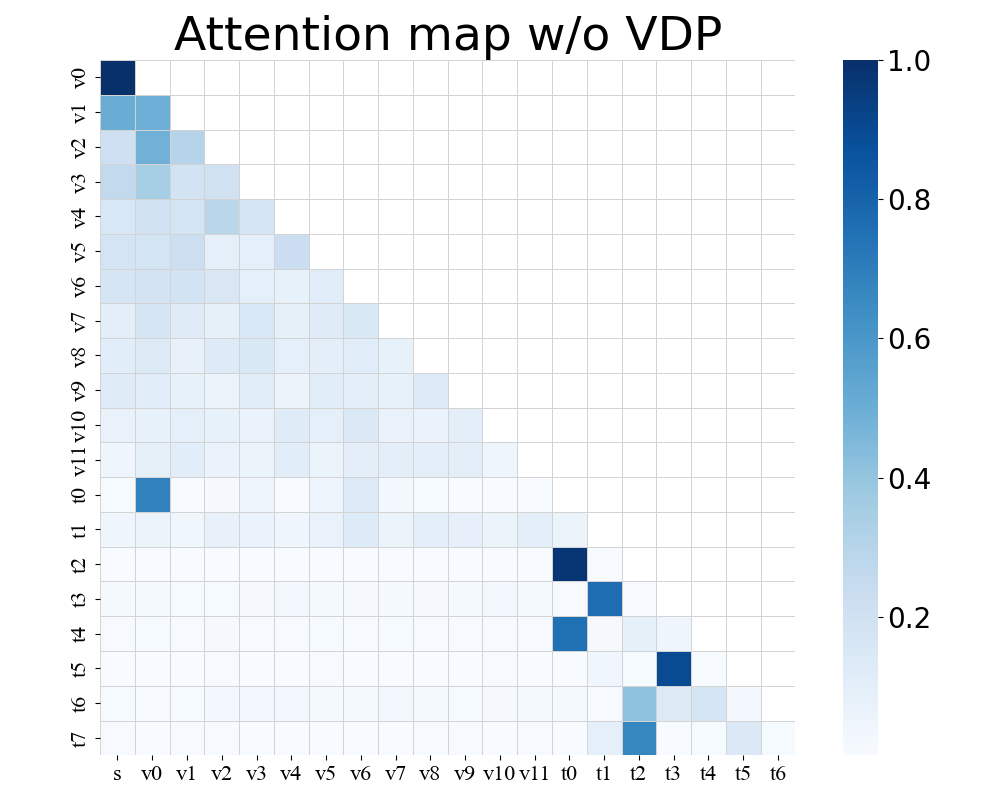}
        \caption{head=31}
     \end{subfigure}
    \caption{Visualization of attention weight in the last layer of Llama \textbf{without}  VPD, showing that the attention is mainly focused on text tokens.}
    \label{fig:attn_mapsWO}
\end{figure*}

\paragraph{Visualization of Attention Maps of Llama}
Here, we provide more attention maps from the last layer of Llama models trained with and without VPD. As shown in Figure~\ref{fig:attn_maps} and \ref{fig:attn_mapsWO}, VPD significantly enhances the model’s ability to prioritize relevant visual features, thereby improving its overall accuracy.

\end{document}